\documentclass{article}
\usepackage[dvipsnames]{xcolor}

\usepackage[accepted]{preamble/icml2020}
\usepackage[utf8]{inputenc} 
\usepackage[T1]{fontenc}    
\usepackage{url}            
\usepackage{booktabs}
\usepackage{amsfonts}       
\usepackage{nicefrac}       
\usepackage{microtype}  
\usepackage{titling}
\usepackage{multirow}
\usepackage{multicol}
\usepackage{amsmath}
\usepackage{bbm}
\usepackage{xspace}
\usepackage{diagbox}
\usepackage{tabularx}
\usepackage{graphicx}
\usepackage{wrapfig}
\usepackage{tikz}
\usepackage{environ}
\usepackage{subcaption}
\usepackage{pgfplots}
\usepackage{hyperref}
\usepackage{amsthm}
\usepackage{todonotes}
\usepackage{xspace}
\usepackage[title]{appendix}
\usepackage{etoolbox}

\usepackage[capitalise, noabbrev]{cleveref}

\BeforeBeginEnvironment{appendices}{\clearpage}

\usepackage{amsmath,amsfonts,bm}

\def\eqref#1{equation~\ref{#1}}

\def\1{\bm{1}}

\def\eps{{\epsilon}}

\DeclareMathAlphabet{\mathsfit}{\encodingdefault}{\sfdefault}{m}{sl}
\SetMathAlphabet{\mathsfit}{bold}{\encodingdefault}{\sfdefault}{bx}{n}

\pgfplotsset{compat=newest}
\pgfplotsset{every axis/.append style={
  label style={font=\LARGE},
  tick label style={font=\Large}  
  }}
\usepgfplotslibrary{groupplots}
\usepgfplotslibrary{dateplot}
\DeclareCaptionSubType*[alph]{table}
\DeclareCaptionSubType*[alph]{figure}
\newtheoremstyle{named}{}{}{\itshape}{}{\bfseries}{.}{.5em}{\thmnote{#3's }#1}
\theoremstyle{named}
\newtheorem{definition}{Definition}
\crefname{appsec}{Appendix}{Appendices}

\makeatletter
\renewcommand{\ALG@name}{Procedure}
\makeatother

\newcommand{\deepobs}{\textsc{DeepOBS}\xspace}

\newcommand{\weightingearly}{\textit{CPE}}

\newcommand{\updaterule}{\mathcal{U}}
\newcommand{\optimizer}{\mathcal{M}}
\newcommand{\prior}{p}
\newcommand{\impulseatk}{$\omega_i = \displaystyle \1_\mathrm{i = K}$}

\newcolumntype{Z}{>{\setbox0=\hbox\bgroup}c<{\egroup}@{\hspace*{-\tabcolsep}}}

\definecolor{mygreen}{RGB}{77,175,74}
\definecolor{myred}{RGB}{228,26,28}
\definecolor{myblue}{RGB}{55,126,184}
\definecolor{myviolet}{RGB}{152,78,163}
\newcommand{\papertitle}{%
Optimizer Benchmarking Needs to Account for Hyperparameter Tuning}
\icmltitlerunning{\papertitle}

\definecolor{color1_app}{rgb}{0.00392156862745098, 0.45098039215686275, 0.6980392156862745}
\definecolor{color2_app}{rgb}{0.8705882352941177, 0.5607843137254902, 0.0196078431372549}
\definecolor{color3_app}{rgb}{0.00784313725490196, 0.6196078431372549, 0.45098039215686275}
\definecolor{color4_app}{rgb}{0.8352941176470589, 0.3686274509803922, 0.0}
\definecolor{color5_app}{rgb}{0.8, 0.47058823529411764, 0.7372549019607844}
\definecolor{color6_app}{rgb}{0.792156862745098, 0.5686274509803921, 0.3803921568627451}
\definecolor{color7_app}{rgb}{0.984313725490196, 0.6862745098039216, 0.8941176470588236}
\definecolor{color8_app}{rgb}{0.5803921568627451, 0.5803921568627451, 0.5803921568627451}
\definecolor{color9_app}{rgb}{0.9254901960784314, 0.8823529411764706, 0.2}
\definecolor{color10_app}{rgb}{0.33725490196078434, 0.7058823529411765, 0.9137254901960784}

\newcommand{\fixed}{$^{C}$}
\newcommand{\Adagrad}{Adagrad\xspace}
\newcommand{\Adam}{Adam\xspace}
\newcommand{\AdamLR}{Adam-LR\xspace}
\newcommand{\AdamDecay}{Adam-W\fixed D}
\newcommand{\SGD}{SGD-LR\xspace}
\newcommand{\SGDM}{SGD-M\xspace}
\newcommand{\SGDMC}{SGD-M\fixed\xspace}
\newcommand{\SGDMW}{SGD-MW\xspace}
\newcommand{\SGDMCWC}{SGD-M\fixed W\fixed\xspace}
\newcommand{\SGDMCWCLRD}{SGD-M\fixed D}
\newcommand{\SGDDecay}{\SGDMCWCLRD}
\newcommand{\SGDLreff}{SGD-LR\textsubscript{eff}\xspace}

\newcommand{\lreff}{\ensuremath{\gamma}\textsuperscript{eff}}

\begin{document}

\twocolumn[
\icmltitle{\papertitle}

\icmlsetsymbol{equal}{*}
\icmlsetsymbol{workatidiap}{\#}

\begin{icmlauthorlist}
\icmlauthor{Prabhu Teja S}{equal,idiap,epfl}
\icmlauthor{Florian Mai}{equal,idiap,epfl}
\icmlauthor{Thijs Vogels}{epfl}
\icmlauthor{Martin Jaggi}{epfl}
\icmlauthor{Fran\c{c}ois Fleuret}{epfl,unige,workatidiap}
\end{icmlauthorlist}

\icmlaffiliation{idiap}{Idiap Research Institute, Switzerland}
\icmlaffiliation{epfl}{EPFL, Switzerland}
\icmlaffiliation{unige}{University of Geneva, Switzerland. \textsuperscript{\#} Work done at Idiap Research Institute}

\icmlcorrespondingauthor{Prabhu Teja S}{prabhu.teja@idiap.ch}
\icmlcorrespondingauthor{Florian Mai}{fmai@idiap.ch}

\icmlkeywords{Optimization, Benchmarking, ICML}

\vskip 0.3in]

\printAffiliationsAndNotice{\icmlEqualContribution} %

\begin{abstract}

    The performance of optimizers, particularly in deep learning, depends considerably on their chosen hyperparameter configuration. The efficacy of optimizers is often studied under near-optimal problem-specific hyperparameters, and finding these settings may be prohibitively costly for practitioners. In this work, we argue that a fair assessment of optimizers' performance must take the computational cost of hyperparameter tuning into account, i.e., how easy it is to find good hyperparameter configurations using an automatic hyperparameter search. Evaluating a variety of optimizers on an extensive set of standard datasets and architectures, our results indicate that Adam is the most practical solution, particularly in low-budget scenarios.
\end{abstract}

\section{Introduction}
\begin{table*}[ht]
  \centering
  \caption{
    Experimental settings shown in the original papers of popular optimizers.
    The large differences in test problems and tuning methods make them difficult to compare.
    $\gamma$ denotes learning rate, $\mu$ denotes momentum, $\lambda$ is the weight decay coefficient.
  }
  \vspace{-0.5\baselineskip}
  \scriptsize
  \label{tab:tune-fab}
  \begin{tabularx}{\textwidth}{X l l l}
    \toprule
    Method                                         & Datasets              & Network architecture   & Parameter tuning methods                                    \\
    \midrule
    SGD with momentum~\citep{pmlr-v28-sutskever13} & Artificial datasets   & Fully-connected        & $\mu=0.9$ for first 1000 updates                            \\
                                                   & MNIST                 & LSTM                   & then $\mu\in\{0, 0.9, 0.98, 0.995\}$.                       \\
                                                   &                       &                        & other schedules for $\mu$ are used                          \\
                                                   &                       &                        & $\&$ $log_{10}(\gamma)\in\{-3, -4, -5, -6\}$                \\
    \cmidrule(lr){1-4}
    Adagrad~\citep{duchi2011adaptive}              & ImageNet ranking      & Single layer           & Perfomance on dev-set                                       \\
                                                   & Reuter RCV1           & Handcrafted features   &                                                             \\
                                                   & MNIST                 & Histogram features                                                                   \\
                                                   & KDD Census            &                                                                                      \\
    \cmidrule(lr){1-4}
    Adam~\citep{kingma2014adam}                    & IMDb                  & Logistic regression    & $\beta_1\in \{0, 0.9\}$                                     \\
                                                   & MNIST                 & Multi-layer perceptron & $\beta_2\in\{0.99, 0.999, 0.9999\}$                         \\
                                                   & CIFAR 10              & Convolutional network  & $log_{10}(\gamma)\in\{-5, -4, -3, -2, -1\}$                 \\
    \cmidrule(lr){1-4}
    AdamW~\citep{loshchilov2018decoupled}          & CIFAR 10              & ResNet CNN             & $log_{2}(\gamma)\in\{-11, -10\dots-1, 0 \}$                 \\
                                                   & ImageNet 32$\times$32 &                        & $log_{2}(\lambda)\in log_{2}(10^{-3})+\{-5, -4, \dots, 4\}$ \\
    \bottomrule
  \end{tabularx}
\end{table*}
\begin{figure}[!b]
    \centering
    \includegraphics[width=\linewidth]{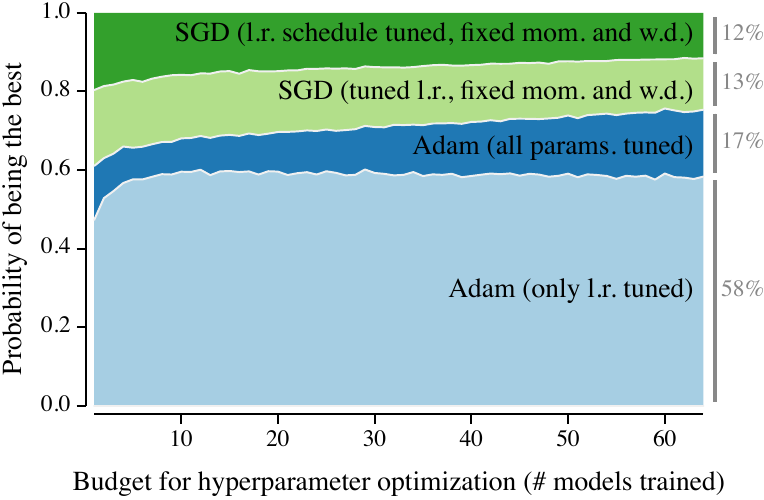}
    \caption{
        Hyperparameter optimization budget affects the performance of optimizers. 
        We show the probability of finding a hyperparameter configuration for an optimizer that performs the best at a given search budget on any task (sampled from our benchmark).  
        This is encoded by the height of the respective area in the chart. 
        Generally, we see that tuning more hyperparameters becomes more useful with higher budgets. On our 9 diverse tasks that include vision problems, natural language processing, regression and classification, tuning only the learning rate for Adam is the most reliable option, even at large budgets.
    }
    \label{fig:prob_plot}
\end{figure}

With the ubiquity of deep learning in various applications, a multitude of first-order stochastic optimizers~\citep{robbins1951stochastic} have been in vogue. They have varying algorithmic components like momentum~\citep{pmlr-v28-sutskever13} and adaptive learning rates~\citep{Tieleman2012,duchi2011adaptive,kingma2014adam}. As the field grows with newer variants being proposed, the standard method to benchmark the performance of these optimizers has been to compare the best possible generalization performance. While it is certainly an important characteristic to be taken into account, we argue that in practice an even more important characteristic is the performance achievable with available resources. A similar view of performance measurement has been recently argued for in the deep learning community owing to the strong debate on sustainable and GreenAI~\citep{strubell2019energy, schwartz2019greenai}.

The performance of optimizers strongly depends on the choice of hyperparameter values such as the learning rate. In the machine learning research community, the sensitivity of models to hyperparameters has been of great debate recently, where in multiple cases, reported model advances did not stand the test of time because they could be explained by better hyperparameter tuning~\citep{DBLP:conf/nips/LucicKMGB18, melis2018on, DBLP:conf/aaai/0002IBPPM18, dodge2019show}. This has led to calls for using automatic hyperparameter optimization methods (HPO) with a fixed budget for a fairer comparison of models~\citep{sculley2018winner's, Feurer2019, DBLP:journals/jair/EggenspergerLH19}. This eliminates biases introduced by humans through manual tuning. For industrial applications, automated machine learning (AutoML, \citealp{DBLP:books/sp/HKV2019}), which has automatic hyperparameter optimization as one of its key concepts, is becoming increasingly more important. In all these cases, an optimization algorithm that achieves good performances with relatively little tuning effort is arguably substantially more useful than an optimization algorithm that achieves top performance, but reaches it only with a lot of careful tuning effort.
Hence, we advocate that benchmarking the performance obtained by an optimizer must not only avoid manual tuning as much as possible, but also has to account for the cost of tuning its hyperparameters to obtain that performance. %

Works that propose optimization techniques show their performance on various tasks as depicted in Table~\ref{tab:tune-fab}. It is apparent that the experimental settings, as well as the network architectures tested, widely vary, hindering a fair comparison. The introduction of benchmarking suites like \deepobs~\citep{schneider2018deepobs} have standardized the architectures tested on, however, this does not fix the problem of selecting the hyperparameters fairly.
Indeed, recent papers studying optimizer performances may employ grid search to select the best values, but the search spaces are still selected on a per-dataset basis, introducing significant human bias~\cite{schneider2018deepobs,wilson2017marginal,shah2018minimum,choi2019empirical}.
Moreover, as only the best obtained performance is reported, it is unclear how a lower search budget would impact the results.
This leads us to the question: how easy is an optimizer to use, i.e. how quickly can an automatic search method find a set of hyperparameters for that optimizer that result in satisfactory performance?

In this paper, we introduce a simple benchmarking procedure for optimizers that addresses the discussed issues.
By evaluating on a wide range of 9 diverse tasks, we contribute to the debate of adaptive vs. non-adaptive optimizers~\citep{wilson2017marginal, shah2018minimum,DBLP:journals/corr/abs-1806-06763, choi2019empirical}. To reach a fair comparison, we experiment with several SGD variants that are often used in practice to reach good performance. Although a well-tuned SGD variant is able to reach the top performance in some cases, our overall results clearly favor Adam~\cite{kingma2014adam}, as shown in Figure~\ref{fig:prob_plot}. %

\section{The Need to Incorporate Hyperparameter Optimization into Benchmarking}\label{sec:whyhpo}
The problem of optimizer benchmarking is two-fold as it needs to take into account
\begin{enumerate}
  \item how difficult it is to find a good hyperparameter configuration for the optimizer,
  \item the absolute performance of the optimizer.
\end{enumerate}

To see why both are needed, consider Figure~\ref{fig:probstillus}, which shows the loss of four different optimizers as a function of their only hyperparameter $\theta$ (by assumption). If we only consider requirement \#1, optimizer C would be considered the best, since every hyperparameter value is the optimum. However, its absolute performance is poor, making it of low practical value. Moreover, due to the same shape, optimizers A and B would be considered equally good, although optimizer A clearly outperforms B. On the other hand, if we only consider requirement \#2, optimizers B and D would be considered equally good, although optimizer D's optimum is harder to find. 

As we discuss in Section~\ref{sec:related-work}, no existing work on optimizer benchmarking takes both requirements into account. Here we present a formulation that does so in Procedure~\ref{alg:benchmark}.
\begin{figure}[htbp]
    \begin{subfigure}[t]{0.47\linewidth}
        \resizebox{\linewidth}{!}{
            \def\offset{-4}
            \begin{tikzpicture}[scale=0.5]
                \draw[thick,->] (0,0) -- (9.5,0);
                \node at (4.75, -.75) {Hyperparameter $\theta$};
                \draw[thick,->] (0,0) -- (0,13 + \offset);
                \node[rotate=90] at (-0.5, 4.5){Expected loss};
                \draw[myblue, very thick] (1,10 + \offset) to[out=0, in=105] (2, 8.5 + \offset);
                \draw[myblue, very thick] plot[smooth, domain=2:6] (\x, \x*\x - 8*\x + 20.5 + \offset);
                \draw[myblue, very thick] (6, 8.5 + \offset) to[out=75, in=180] (7, 10 + \offset)
                node[right] {\footnotesize Optimizer A};
                \draw[myviolet, very thick] (1,12 + \offset) to[out=0, in=100] (3, 10.5 + \offset);
                \draw[myviolet, very thick] plot[smooth, domain=3:5] (\x, 4*\x*\x - 32*\x + 70.5 + \offset);
                \draw[myviolet, very thick] (5, 10.5 + \offset) to[out=90, in=180] (7, 12 + \offset)
                node[right]{\footnotesize Optimizer D};
                \draw[myred, very thick] (1,12 + \offset) to[out=0, in=105] (2, 10.5 + \offset);
                \draw[myred, very thick] plot[smooth, domain=2:6] (\x, \x*\x - 8*\x + 22.5 + \offset);
                \draw[myred, very thick] (6, 10.5 + \offset) to[out=75, in=180] (7, 12 + \offset)
                node[above right] {\footnotesize Optimizer  B};
                \draw[mygreen, very thick] plot[domain=1:7](\x, 12.5 + \offset)%
                node[above right]{\footnotesize Optimizer C};
            \end{tikzpicture}}
        \caption{Illustration. It is important to consider both the absolute performance of optimizers as well as the tuning effort to get to good performances.}
        \label{fig:probstillus}
    \end{subfigure}
    \hspace{0.3cm}
    \begin{subfigure}[t]{0.47\linewidth}
        \resizebox{\textwidth}{!}{
            \begin{tikzpicture}[scale=0.5]
                \draw[thick,->] (0,0) -- (9.5,0);
                \node at (4.75, -.75) {Hyperparameter $\theta$};
                \draw[thick,->] (0,0) -- (0,9);
                \node[rotate=90] at (-0.5, 4.5){Expected loss $\mathcal{L}(\theta)$};
                \begin{scope}[yshift=1cm]
                    \draw[RoyalPurple, very thick] (1,5) ..controls (3, 2) .. (7, 4) node[right] {\footnotesize Optimizer F};
                    \draw[magenta, very thick] (1, 5.5) .. controls (4, 5) and (4, -1) .. (4.5, 3) to[out=85, in=190] (7, 4.5) node[right]{\footnotesize Optimizer E};
                    \node[RoyalPurple,label={270:\color{RoyalPurple}{$\theta_F^\star$}},circle,fill,inner sep=2pt] at (3.25, 2.65) {};
                    \node[magenta, label={270:\color{magenta}{$\theta_E^\star$}},circle,fill,inner sep=2pt] at (4.15, 1.65) {};
                \end{scope}
            \end{tikzpicture}}
        \caption{Illustration. While optimizer E can achieve the best performance after careful tuning, optimizer F is likely to provide better performance under a constrained HPO budget.}
        \label{fig:optimillus}
    \end{subfigure}
\end{figure}

We have already established that fairly comparing optimizers needs to account for how easy it is to find good hyperparameter values.
When proposing new optimization methods, most often algorithm designers only specify the permissible set of values the hyperparameters can take, and informally provide some intuition of good values.
For example, for Adam, \citet{kingma2014adam} bound $\beta_1, \beta_2$ to $[0, 1)$ and specify that they should be close to 1.
These are valuable information for users of their algorithm, but they do not allow to formally incorporate that information into a benchmarking procedure.
Instead, we argue that we need to redefine what constitutes an optimizer in such a way that prior knowledge over reasonable hyperparameter values is included.

\begin{definition}
  An optimizer is a pair $\optimizer = (\updaterule_{\Theta}, \prior_{\Theta})$, which applies its update rule $\updaterule(S_{t}; \Theta)$ at each step $t$ depending on its current state $S_t$. It is parameterized through N hyperparameters $\Theta = (\theta_1,\dots,\theta_N)$ with respective permissible values $\theta_i \in H_i \,\,\forall i$, and 
  $\prior_{\Theta} : (\Theta \rightarrow \mathbb{R}$) defines a probability distribution over the hyperparameters.
  
\end{definition}
  In the example above, we could describe the intuition that $\beta_1, \beta_2$ should be close to 1 by the random variables $\hat{\beta_{1}} = 1 - 10^{c_1}, \hat{\beta_{2}} = 1 - 10^{c_2}$, where $c_1, c_2 \sim U(-10, -1)$.

Let $\mathcal{L}(\Theta_1)$ refer to the performance (say, test loss) of $\optimizer$ with the specific hyperparameter choice $\Theta_1$.

\begin{algorithm}[tb]
   \caption{Benchmark with `expected quality at budget'}
   \label{alg:benchmark}
\begin{algorithmic}
   \STATE {\bfseries Input:} Optimizer $O$, cross-task hyperparameter prior $\Theta_O$, task $T$, tuning budget $B$
   \STATE {\bfseries Initialization:} Pre-compute a \emph{library} of size $\gg\hspace{-4pt}B$ with validation losses achieved on task $T$ with optimizer $O$ using hyper-parameters sampled from $\Theta_O$.
   \STATE Initialize $list \gets [\,]$.
   \FOR{$R$ repetitions}
   \STATE Simulate hyperparameter search with budget $B$:
    \STATE --\hspace{6pt} $S\gets$ sample $B$ elements from \emph{library}.
    \STATE --\hspace{6pt} $list\gets [\textsc{best}(S),\; \ldots list]$.
   \ENDFOR
   \STATE {\bfseries return} \textsc{mean}($list$), or other statistics
\end{algorithmic}
\end{algorithm}

Let us assume that there are two optimizers E \& F, both with a single hyperparameter $\theta$, but no prior knowledge of particularly good values, i.e., the prior is a uniform distribution over the permissible range. Let their loss surface be $\mathcal{L}_E$ and $\mathcal{L}_F$, respectively. As Figure~\ref{fig:optimillus} shows, the minimum of $\mathcal{L}_E$ is lower than that of $\mathcal{L}_F$ (denoted by $\theta_E^\star$ and $\theta_F^\star$) i.e. $\mathcal{L}_E(\theta_E^\star)< \mathcal{L}_F(\theta_F^\star)$. However, the minimum of $\mathcal{L}_E$ is much sharper than that of $\mathcal{L}_F$, and in most regions of the parameter space F performs much better than E. This makes it easier to find configurations that perform well. This makes optimizer-F an attractive option when we have no prior knowledge of the good parameter settings. Previous benchmarking strategies compare only $\theta_E^\star$ and $\theta_F^\star$. It is obvious that in practice, optimizer-F may be an attractive option, as it gives `good-enough' performance without the need for a larger tuning budget.

We incorporate the relevant characteristics of the hyperparameter optimization surface described above into benchmarking through Procedure~\ref{alg:benchmark}. In the proposed protocol, we use Random Search~\citep{bergstra2012random} with the optimizers' prior distribution to search the hyperparameter space.
The quality of the optimizers can then be assessed by inspecting the maximum performance attained after $k$ trials of random search.
However, due to the stochasticity involved in random search, we would usually have to repeat the process many times to obtain a reliable estimate of the distribution of performances after budget $k$. 
We instead use the bootstrap method~\citep{tibshirani1993introduction} that re-samples from the empirical distribution (termed $library$ in Procedure~\ref{alg:benchmark}). When we need the mean and variance of the best attained performance after budget $k$, we use the method proposed by \citet{dodge2019show} to compute them exactly in closed form. We provide the details of the computation in \cref{app:expmax}.

Our evaluation protocol has distinct advantages over previous benchmarking methods that tried to incorporate automatic hyperparameter optimization methods.
First, our evaluation protocol is entirely free of arbitrary human choices that bias benchmarking: The only free parameters of random search itself are the search space priors, which we view as part of the optimizer.
Secondly, since we measure and report the performance of Random Search with low budgets, we implicitly characterize the loss surface of the hyperparameters: In terms of Figure~\ref{fig:optimillus}, optimizer-F with its wide minimum will show good performance with low budgets, whereas optimizer-E can be expected to show better performance with high budgets. Such characterizations would not be possible if one only considered the performance after exhausting the full budget.
Finally, our evaluation protocol allows practitioners to choose the right optimizer for their budget scenarios.

\paragraph{Discussion of alternative choices}
In theory, our general methodology could also be applied with a different hyperparameter optimization technique that makes use of prior distributions, e.g., drawing the set of initial observations in Bayesian methods.
However, those usually have additional hyperparameters, which can act as potential sources of bias. 
Moreover, the bootstrap method is not applicable when the hyperparameter trials are drawn dependently, and repeating the hyperparameter optimization many times is practically infeasible.

In our protocol we consider the number of random search trials as the unit of budget, and not computation time. This is done so as to not violate the independence assumption in the method by \citet{dodge2019show} in \cref{app:expmax}. We, empirically, show in \cref{sec:time-budget} that the conclusions of this paper are still valid when time is used as the unit of budget as well.

\section{Related Work}
\label{sec:related-work}
Benchmarking of optimizers is a relatively unstudied subject in literature. \citet{schneider2018deepobs} recently released a benchmark suite for optimizers that evaluates their peak performance and speed, and the performance measure is assessed as the sensitivity of the performance to changes of the learning rate. Our work primarily takes its genesis from the study by \citet{wilson2017marginal} that finds SGD-based methods as easy to tune as adaptive gradient methods. They perform grid earch on manually chosen grids for various problems and conclude that both SGD and Adam require similar grid search effort. However, their study lacks a clear definition of what it means to be tunable (easy-to-use) and tunes the algorithms on manually selected, dataset dependent grid values. The study by \citet{shah2018minimum} applies a similar methodology and comes to similar conclusions regarding performance. Since both studies only consider the best parameter configuration, their approaches cannot quantify the efforts expended to find the hyperparameter configuration that gives the best setting; they would be unable to identify the difference between optimizer among B and D in Figure~\ref{fig:probstillus}. In contrast, the methodology in our study is able to distinguish all the cases depicted in Figure~\ref{fig:probstillus}.

There exist few works that have tried to quantify the impact of hyperparameter setting in ML algorithms. %
For decision tree models, \citet{mantovani2018empirical} count the number of times the tuned hyperparameter values are (statistically significantly) better than the default values.
\citet{probstTunabilityImportanceHyperparameters} define tunability of an ML algorithm as the performance difference between a reference configuration (e.g., the default hyperparameters of the algorithm) and the best possible configuration on each dataset. This metric is comparable across ML algorithms, but it disregards entirely the absolute performance of ML algorithms; thereby being unable to differentiate between optimizers B and D in \cref{fig:probstillus}.

In a concurrent study, \citet{choi2019empirical} show that there exists a hierarchy among optimizers such that some can be viewed as specific cases of others and thus, the general optimizer should never under-perform the special case (with appropriate hyperparameter settings). Like in our study, they suggest that the performance comparison of optimizers is strongly predicated on the hyperparameter tuning protocol. 
However, their focus is on the best possible performance achievable by an optimizer and does not take into account the tuning process. Also, the presence of a hierarchy of optimizers does not indicate how easy it is to arrive at the hyperparameter settings that help improve the performance of the more \emph{general} optimizer. Moreover, while the authors claim their search protocol to be relevant for practitioners, the search spaces are manually chosen \emph{per dataset}, constituting a significant departure from a realistic AutoML scenario considered in our paper. Since, the focus is only on the best attainable performance, it construed as being benchmarking theoretically infinite budget scenarios.

In a recent work, \citet{dodge2019show} propose to use the performance on the validation set along with the test set performance. They note that the performance conclusions reached by previously established NLP models differ widely from the published works when additional hyperparameter tuning budget is considered. They recommend a checklist to report for scientific publications that includes details of compute infrastructure, runtime, and more importantly the hyperparameter settings used to arrive at those results like bounds for each hyperparameter, HPO budget and tuning protocols. They recommend using expected validation performance at a given HPO budget as a metric, along with the test performance.

There has been recent interest in optimizers that are provably robust to hyperparameter choices, termed the \textsc{aProx} family~\citep{Asi201908018, asi2019stochastic}. \citeauthor{Asi201908018} experimentally find that, training a Residual network~\citep{he2016deep} on CIFAR-10, SGD converges only for a small range of initial learning rate choices, whereas Adam exhibits better robustness to learning rate choices; their findings are in line with our experiments that it is indeed easier to find good hyperparameter configurations for Adam.

\citet{metz2020using} propose a large range of tasks, and propose to collate hyperparameter configurations over those. They show that the optimizer settings thus collated, that are problem agnostic like us, generalize well to unseen tasks too. 
\section{Optimizers and Their Hyperparameters}
In \cref{sec:whyhpo}, we argued that an optimizer is a combination of update equation, and the probabilistic prior on the search space of the hyperparameter values. Since we are considering a setup akin to AutoML with as little human intervention as possible, these priors have to be independent of the dataset.
As we view the hyperparameter priors as a part of the optimizer itself, we argue that they should be prescribed by algorithm designers themselves in the future. However, in the absence of such prescriptions for optimizers like Adam and SGD, we provide a simple method to estimate suitable priors in \cref{sec:calibration}.

\subsection{Parameters of the Optimizers}
To compare the tunability of adaptive gradient methods to non-adaptive methods, we chose the most commonly used optimizers from both the strata; SGD and SGD with momentum for non-adaptive methods, and Adagrad and Adam for adaptive gradient methods. 
Since adaptive gradient methods are said to work well with their default hyperparameter values already, we additionally employ a default version of Adam where we only tune the initial learning rate and set the other hyperparameters to the values recommended in the original paper~\citep{kingma2014adam} (termed \AdamLR). Such a scheme has been used by \citeauthor{schneider2018deepobs} too. A similar argument can be made for SGD with momentum (termed \SGDM): thus we experiment with a fixed momentum value of 0.9 (termed \SGDMC), which we found to be the most common momentum value to lead to good performance during the calibration phase. 

In addition to standard parameters in all optimizers, we consider weight decay with SGD too. SGD with weight decay can be considered as an optimizer with two steps where the first step is to scale current weights with the decay value, followed by a normal descent step~\citep{loshchilov2018decoupled}. Therefore, we conduct two additional experiments for SGD with weight-decay: one where we tune weight-decay along with momentum (termed \SGDMW), and one where we fix it to $10^{-5}$ (termed \SGDMCWC) along with the momentum being fixed to 0.9, which again is the value for weight decay we found to be the best during calibration. We incorporate a ``Poly'' learning rate decay scheduler ($\gamma_t = \gamma_0 \times (1-\frac{t}{T})^p$)~\citep{liu2016parsenet} for \SGDMCWC (termed \SGDDecay). This adds only one tunable hyperparameter (exponent $p$). We also experimented with Adam with learning rate decay scheduler (termed \AdamDecay), but reserve this discussion for the \cref{app:fullplots}, as it did not yield sizeable improvements over \AdamLR or \Adam in the problems tested. The full list of optimizers we consider is provided in Table~\ref{tab:optim_tunable}, out of which we discuss \AdamLR, \Adam, \SGDMCWC, \SGDMW, and \SGDDecay~in the main paper. The rest of them are presented in \cref{app:fullplots}. 

Manually defining a specific number of epochs can be biased towards one optimizer, as one optimizer may reach good performance in the early epochs of a single run, another may reach higher peaks more slowly. In order to alleviate this, it would be possible to add the number of training epochs as an additional hyperparameter to be searched.  Since this would incur even higher computational cost, we instead use validation set performance as stopping criterion. Thus we stop training when the validation loss plateaus for more than~2 epochs or if the number of epochs exceeds the predetermined maximum number as set in \deepobs. %

\subsection{Calibration of Hyperparameter Prior Distributions}\label{sec:calibration}
As mentioned previously, we use random search for optimizing the hyperparameters, which requires distributions of random variables to sample from. Choosing poor distributions to sample from impacts the performance, resulting in unfair comparisions, and may break requisite properties (e.g. learning rate is non-negative). For some of the parameters listed in Table~\ref{tab:optim_params}, obvious bounds exist due their mathematical properties, or have been prescribed by the optimizer designers themselves. For example, \citet{kingma2014adam} bound $\beta_1, \beta_2$ to $[0, 1)$ and specify that they are close to 1. In the absence of such prior knowledge, we devise a simple method to determine the priors.

We use Random Search on a large range of admissible values on each task specified in \deepobs to obtain an initial set of results.
We then retain the hyperparameters which resulted in performance within $20\%$ of the best result obtained. For each of the hyperparameters in this set, we fit the distributions in the third column of Table~\ref{tab:optim_params} using maximum likelihood estimation. Several recent works argue that there exists a complex interplay between the hyperparameters~\citep{smith2018dont,shallue2019measuring}, but we did not find modelling these to be helpful (\cref{app:sec:lreff}). Instead, we make a simplifying assumption that all the hyperparameters can be sampled independent of each other. We argue that these distributions are appropriate; the only condition on learning rate is non-negativity that is inherent to the log-normal distribution, momentum is non-negative with a usual upper bound of 1, $\beta$s in Adam have been prescribed to be less than 1 but close to it, $\epsilon$ is used to avoid division by zero and thus is a small positive value close to 0. We did not include $p$ of the learning rate decay schedule in the calibration step due to computational constraints, and chose a fixed plausible range such that the value used by \cite{liu2016parsenet} is included. We report the parameters of the distributions obtained after the fitting in Table~\ref{tab:optim_params}. The calibration step is not included in computing the final performance scores, as the calibrated priors are re-usable across tasks and datasets.

\begin{table}[ht]
    \caption{
    Optimizers evaluated. For each hyperparameter, we calibrated a `prior distribution' to give good results across tasks (Section~\ref{sec:calibration}). 
    $\mathcal{U}[a, b]$ is the continuous uniform distribution on $[a, b]$. 
    Log-uniform($a$, $b$) is a distribution whose logarithm is $\mathcal{U}[a, b]$.
    Log-normal($\mu$,$\sigma$) is a distribution whose logarithm is $\mathcal{N}(\mu, \sigma^2)$%
    }
    \label{tab:optim_params}
    \centering
    \small
    \begin{tabularx}{\linewidth}{llX}
        \toprule
        Optimizer                                    & Tunable parameters & Cross-task prior    \\ \midrule
        SGD                  & Learning rate      & Log-normal(-2.09, 1.312) \\
                                                     & Momentum           & $\mathcal{U}{[}0, 1{]}$  \\
                                                     & Weight decay       & Log-uniform(-5, -1)      \\ 
                                                     & Poly decay  ($p$)  & $\mathcal{U}{[}0.5, 5{]}$  \\\cmidrule(lr){1-3}
        Adagrad                                      & Learning rate      & Log-normal(-2.004, 1.20) \\ \cmidrule(lr){1-3}
        Adam                                         & Learning rate      & Log-normal(-2.69, 1.42)  \\
                                                     & $\beta_1, \beta_2$ & 1 - Log-uniform(-5, -1)      \\
                                                     & $\eps$             & Log-uniform(-8, 0)       \\ \bottomrule
    \end{tabularx}
\end{table}

\section{Experiments and Results}\label{sec:experiments}
To assess the performance of optimizers for the training of deep neural networks, we benchmark using the open-source suite \deepobs \citep{schneider2018deepobs}. The architectures and datasets we experiment with are given in Table~\ref{tab:exp_archs}. We refer the reader to \citet{schneider2018deepobs} for specific details of the architectures. To obtain a better balance between vision and NLP applications, we added an LSTM network with the task of sentiment classification in the IMDB dataset~\citep{maas2011learning}, details of which are provided in Appendix~\ref{app:architectures}.

\begin{table}[ht]
  \centering
  \caption{
    Models and datasets used.
    We use the DeepOBS benchmark set~\citep{schneider2018deepobs}.
    Details are provided in Appendix~\ref{app:architectures}.
  }
  \vspace{-0.5\baselineskip}
  \label{tab:exp_archs}
  \small
  \begin{tabularx}{\linewidth}{lX}
    \toprule
    Architecture            & Datasets                \\ \midrule
    Convolutional net       & FMNIST, CIFAR10/100     \\
    Variational autoencoder & FMNIST, MNIST           \\
    Wide residual network   & SVHN                    \\
    Character RNN           & Tolstoi's War and Peace \\
    Quadratic function      & Artificial datatset     \\ %
    LSTM                    & IMDB                    \\\bottomrule
  \end{tabularx}
\end{table}
\begin{table}[!b]
  \caption{
    Optimizers and tunable parameters.
    SGD($\gamma, \mu, \lambda$) is SGD with $\gamma$ learning rate, $\mu$ momentum, $\lambda$ weight decay coefficient.
    Adagrad($\gamma$) is Adagrad with $\gamma$ learning rate,
    Adam($\gamma, \beta_1, \beta_2, \eps$) is Adam with learning rate $\gamma$,
  }
  \vspace{-0.5\baselineskip}
  \label{tab:optim_tunable}
  \small
  \begin{tabularx}{\linewidth}{lX}
    \toprule
    Optimizer label                                                  & Tunable parameters                                                                         \\ \midrule
    \SGD                                                              & SGD($\gamma, \mu\!\!=\!\!0, \lambda\!\!=\!\!0$)                                            \\
    \SGDM                                                             & SGD($\gamma, \mu, \lambda\!\!=\!\!0$)                                                      \\
    \SGDMC                                          & SGD($\gamma, \mu\!\!=\!\!0.9, \lambda\!\!=\!\!0$)                                          \\
    \SGDMCWC                      & SGD($\gamma, \mu\!\!=\!\!0.9, \lambda\!\!=\!\!10^{-5}$)                                    \\
    \SGDMCWCLRD & SGD($\gamma, \mu\!\!=\!\!0.9, \lambda\!\!=\!\!10^{-5}$) + Poly Decay($p$)             \\
    \SGDMW                                                            & SGD($\gamma, \mu, \lambda$)                                                                \\ \cmidrule(lr){1-2}
    \Adagrad                                                          & Adagrad($\gamma$)                                                                          \\
    \AdamLR                                                           & Adam($\gamma$, $\beta_1\!\!=\!\!0.9$, $\beta_2\!\!=\!\!0.999$, $\epsilon\!\!=\!\!10^{-8}$) \\
    \Adam                                                             & Adam($\gamma, \beta_1, \beta_2, \epsilon$)                                                 \\
    \AdamDecay                                                        & \AdamLR +  Poly Decay($p$) \\
    \bottomrule
  \end{tabularx}
\end{table}

We aim at answering two main questions with our experiments:
First, we look at the performance of various optimizers examined. Related to this, we investigate what effect the number of hyperparameters being tuned has on the performance at various budgets (\cref{sec:res-hyperparams}). 
Second, we consider a problem typically faced in an AutoML scenario: If no knowledge is available a priori of the problem at hand, but only the tuning budget, which optimizer should we use (\cref{sec:summary_stats})?

\subsection{When to Tune More Hyperparameters}\label{sec:res-hyperparams}
To answer the question at which budgets tuning more hyperparameters is preferable, we compare \AdamLR to \Adam, and \SGDMW to \SGDMCWC (\cref{tab:optim_tunable}).
To this end, we show performance for increasing budgets $K$ in \cref{fig:perf_plots}. 
Plots for the other optimizers and budgets are given in \cref{fig:perf_plots_app_full}. 

On all classification tasks, \AdamLR and \SGDMCWC obtain higher performances on average than \Adam and \SGDMW, respectively, till the budget of 16.
Moreover, the first quartile is often substantially lower for the optimizers with many hyperparameters.
For higher budgets, both outperform their counterparts on CIFAR-100 and FMNIST on average and in the second quartile, and \Adam outperforms \AdamLR on IMDB as well.
However, even for the largest budgets, \Adam's first quartile is far lower than \AdamLR's.

On the regression tasks, tuning more hyperparameters only helps for \SGDMW on MNIST-VAE. In all other cases, tuning additional hyperparameters degrades the performance for small budgets, and achieves similar performance at high budgets.

\subsection{Summarizing across datasets} \label{sec:summary_stats}
\begin{figure}[htpb]
  \begin{center}
    \includegraphics[width=0.40\textwidth]{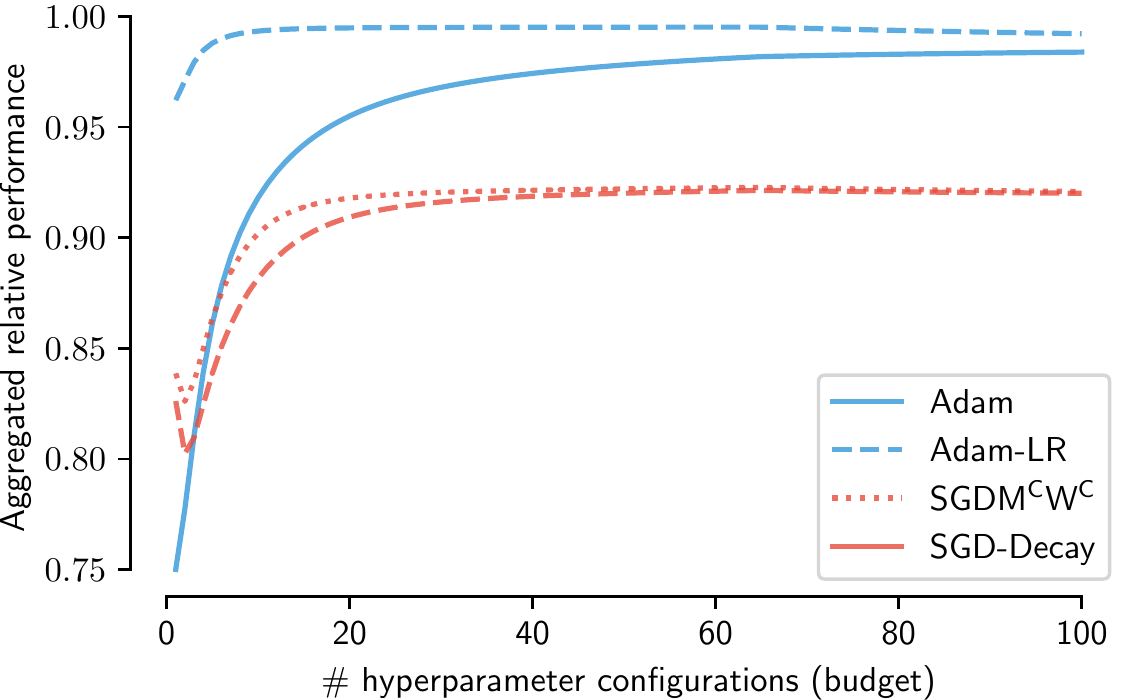}
  \end{center}
  \vspace{-0.5\baselineskip}
  \caption{
    Aggregated relative performance of various optimizers across datasets
  }
  \label{fig:sumstat_}
\end{figure}
We, now, turn to the question of how our choice of optimizer would change in a setting where nothing is known about the problem, \`a la AutoML. %
To this end, we summarize performances across all datasets. 
First, for a given budget, we compute the probability that an optimizer outperforms the others on a randomly chosen task. 
In Figure~\ref{fig:prob_plot}, we compare \Adam, \AdamLR, \SGDMCWC, and \SGDDecay, because we found them to yield the overall best results. 
First, the results reflect the findings from Section~\ref{sec:res-hyperparams} in that tuning more hyperparameters (\Adam) becomes a better relative option the more budget is available. 
However, throughout all tuning budget scenarios, \AdamLR remains by far the most probable to yield the best results.

Figure~\ref{fig:prob_plot} shows that \AdamLR is the most likely to get the best results.
However, it does not show the margin by which the SGD variants underperform. 
To address this issue, we compute summary statistics for an optimizer $o$'s performance after $k$ iterations in the following way:
\[S(o, k) = \frac{1}{|\mathcal{P}|} \sum\limits_{p \in \mathcal{P}}\frac{o(k,p)}{\max\limits_{o' \in \mathcal{O}} o'(k,p)},\]
where $o(k,p)$ denotes the expected performance of optimizer $o \in \mathcal{O}$ on test problem $p \in \mathcal{P}$ after $k$ iterations of hyperparameter search.
In other words, we compute the average relative performance of an optimizer to the best performance of any optimizer on the respective task, at budget~$k$. 

The results are in Figure~\ref{fig:sumstat_} which show that \AdamLR performs very close to the best optimizer for all budgets. 
In early stages of HPO, the SGD variants perform 20\% worse than \AdamLR. This gap narrows to 10\% as tuning budgets increase, but the flatness of the curves for high budgets suggest that they are unlikely to improve further with higher budgets.
\Adam on the other hand steadily improves relative to \AdamLR, and only leaves a 2-3\% gap at high budgets.

\definecolor{color1}{rgb}{0.00392156862745098, 0.45098039215686275, 0.6980392156862745}
\definecolor{color2}{rgb}{0.8705882352941177, 0.5607843137254902, 0.0196078431372549}
\definecolor{color3}{rgb}{0.00784313725490196, 0.6196078431372549, 0.45098039215686275}
\definecolor{color4}{rgb}{0.8352941176470589, 0.3686274509803922, 0.0}
\definecolor{color5}{rgb}{0.8, 0.47058823529411764, 0.7372549019607844}
\definecolor{color6}{rgb}{0.792156862745098, 0.5686274509803921, 0.3803921568627451}
\definecolor{color7}{rgb}{0.984313725490196, 0.6862745098039216, 0.8941176470588236}
\definecolor{color8}{rgb}{0.5803921568627451, 0.5803921568627451, 0.5803921568627451}
\definecolor{color9}{rgb}{0.9254901960784314, 0.8823529411764706, 0.2}
\definecolor{color10}{rgb}{0.33725490196078434, 0.7058823529411765, 0.9137254901960784}

\newcommand{\showcolor}[2]{\textbf{\color{#1}{#2}}}

\begin{figure*}[htp]
    \includegraphics[width=\textwidth]{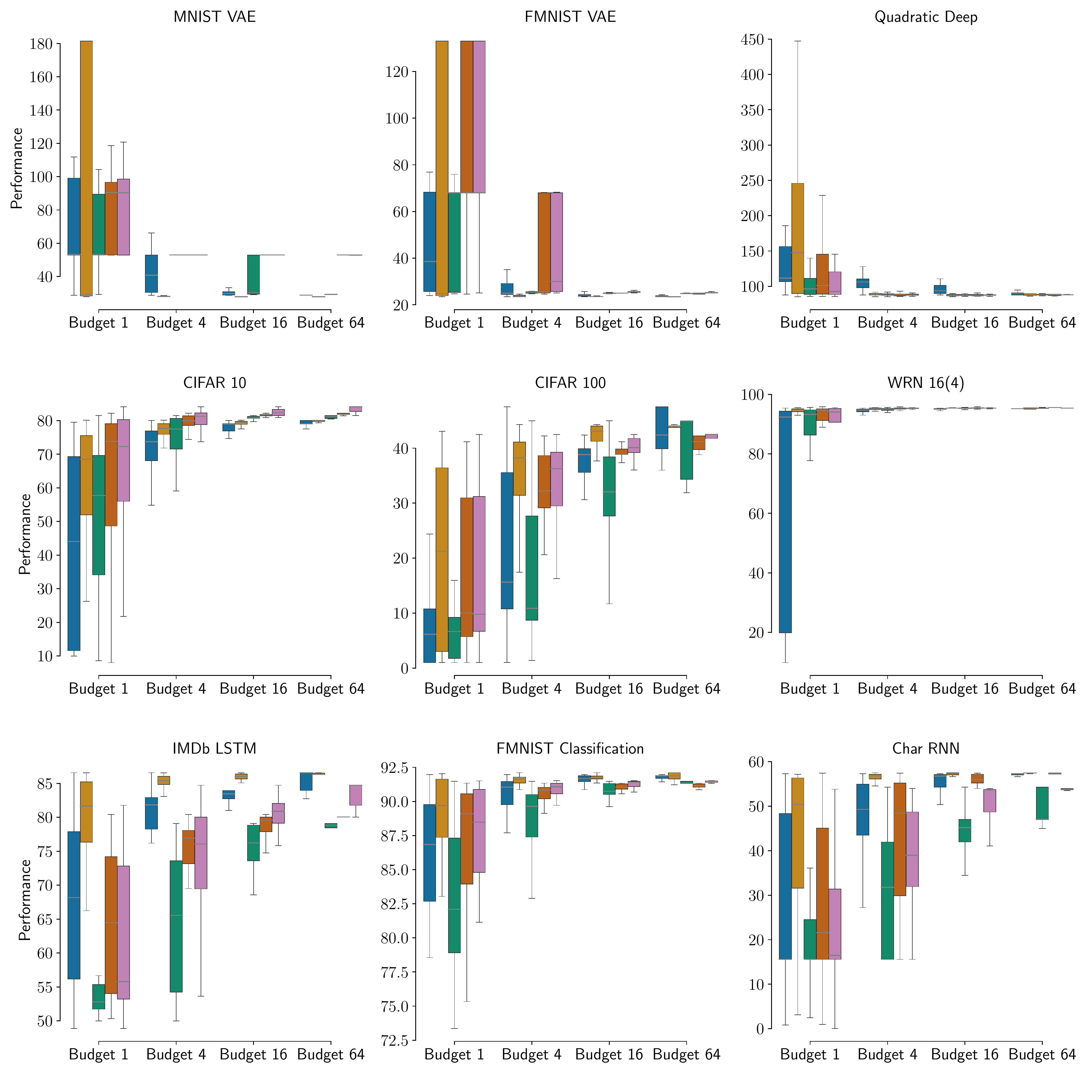}
    \caption{Performance of \showcolor{color1}{\AdamLR}, \showcolor{color2}{\Adam}, \showcolor{color3}{\SGDMCWC}, \showcolor{color4}{\SGDMW}, \showcolor{color5}{\SGDDecay} at various hyperparameter search budgets. Image is best viewed in color. Some of the plots have been truncated to increase readability.}
    \label{fig:perf_plots}
\end{figure*}

\section{Discussion}
The key results of our experiments are two-fold. First, they support the hypothesis that adaptive gradient methods are easier to tune than non-adaptive methods: 
In a setting with low budget for hyperparameter tuning, tuning only Adam's learning rate is likely to be a very good choice; it doesn't guarantee the best possible performance, but it is evidently the easiest to find well-performing hyperparameter configurations for.
While SGD (variants) yields the best performance in some cases, its best configuration is tedious to find, and Adam often performs very close to it.
Hence, in terms of Figure~\ref{fig:optimillus}, SGD seems to be a hyperparameter surface with narrow minima, akin to optimizer E, whereas the minima of Adam are relatively wide, akin to optimizer F.
We investigate the empirical hyperparameter surfaces in Appendix~\ref{sec:app-surfaces} to confirm our hypothesis.
We, thus, state that the substantial value of the adaptive gradient methods, specifically Adam, is its amenability to hyperparameter search. 
This is in contrast to the findings of \citet{wilson2017marginal} who observe no advantage in tunabilty for adaptive gradient methods, and thus deem them to be of `marginal value'.
This discrepancy is explained by the fact that our evaluation protocol is almost entirely free of possible human bias: 
In contrast to them, we do not only avoid manually tuning the hyperparameters through the use of automatic hyperparameter optimization, we also automatically determine the HPO's own hyperparameters by estimating the distributions over search spaces.

Secondly, we find that tuning optimizers' hyperparameters apart from the learning rate becomes more useful as the available tuning budget goes up.
In particular, we find that \Adam approaches the performance of \AdamLR for large budgets.
This is, of course, an expected result, and in line with recent work by \citet{choi2019empirical}, who argue that, with sufficient hyperparameter tuning, a more general optimizer (\Adam) should never under-perform any particular instantiation thereof (\AdamLR).
\citet{choi2019empirical} claim that this point is already reached in `realistic' experiments.
However, in their experiments, \citet{choi2019empirical} tune the search spaces for each problem they consider, thereby assuming apriori knowledge of what constitutes meaningful hyperparameter settings for that specific problem. 
Our results, which are obtained with a protocol that is arguably less driven by human bias, tell a different story: Even with relatively large tuning budget, tuning only the learning rate of Adam is arguably the safer choice, as it achieves good results with high probability, whereas tuning all hyperparameters can also result in a better performance albeit with high variance. 
These observations suggest that optimizers with many tunable hyperparameters have a hyperparameter surface that is less smooth, and that is the reason why fixing e.g. the momentum and weight decay parameters to prescribed 'recipe' values is beneficial in low-resource scenarios.

Our study is certainly not exhaustive: We do not study the effect of a different HPO like a Bayesian HPO on the results, due to prohibitively high computational cost it incurs.
By choosing uni-variate distribution families for the hyperparameters to estimate the priors, we do not account for complex relationships between parameters that might exist. We explore this in \cref{app:sec:lreff} where we use the notion of `effective learning rate'~\citep{shallue2019measuring}, and we find that it helps improve the performance in the lower budgets of hyperparameter optimization. We attribute this to the fact that SGDElrW is effective at  exploiting historically successful $(\gamma, \mu)$ pairs. However, the literature does not provide methods to incorporate these into a probabilistic model that incorporates the causal relationships between them. 

In the future, we suggest that optimizer designers not only study the efficacy and convergence properties, but also provide priors to sample hyperparameters from. Our study demonstrates this to be a key component in determining an optimizer's practical value.

\section{Conclusion}
We propose to include the process of hyperparameter optimization in optimizer benchmarking. In addition to showing peak performance, this showcases the optimizer's ease-of-use in practical scenarios.
We hope that this paper encourages other researchers to conduct future studies on the performance of optimizers from a more holistic perspective, where the cost of the hyperparameter search is included.

\ifdefined\isaccepted
\subsubsection*{Acknowledgments}
Prabhu Teja was supported by the ``Swiss Center for Drones and Robotics - SCDR'' of the Department of Defence, Civil Protection and Sport via armasuisse S+T under project n\textsuperscript{o}050-38.
Florian Mai was supported by the Swiss National Science Foundation under the project Learning Representations of Abstraction for Opinion Summarisation (LAOS), grant number ``FNS-30216''.
The authors thank Frank Schneider and Aaron Bahde for giving us access to the pre-release PyTorch version of \deepobs.
\fi

\bibliography{project_report,OptMLProject}
\bibliographystyle{preamble/icml2020}

\begin{appendices}
\onecolumn
    \icmltitle{{\small Appendix for:} \\ \papertitle}
    \icmlsetsymbol{equal}{*}
    \begin{icmlauthorlist}
    \icmlauthor{Prabhu Teja S}{equal}
    \icmlauthor{Florian Mai}{equal}
    \icmlauthor{Thijs Vogels}{}
    \icmlauthor{Martin Jaggi}{}
    \icmlauthor{Fran\c{c}ois Fleuret}{}
    \end{icmlauthorlist}
    
    \vskip 0.3in    
\newcommand{\conv}{\text{Conv2D}}
\newcommand{\relu}{\text{ReLU}}
\newcommand{\pool}{\text{MaxPool2D}}
\newcommand{\lin}{\text{Linear}}
\newcommand{\drop}{\text{Dropout}}
\newcommand{\smax}{\text{Softmax}}
\newcommand{\ta}[1]{\text{#1}}

\newcommand{\horizfigresize}[2][]{%
    \begin{minipage}{0.48\linewidth}\subfloat[#1]{
            \resizebox{\textwidth}{!}
            {#2}}\end{minipage}}

\section{Architectures of the Models Used in Experiments}\label{app:architectures}
\crefalias{section}{appsec}

Along with the architectures examined by \cite{schneider2018deepobs}, we experiment with an additional network and dataset. We included an additional network into our experimental setup, as \deepobs does not contain a word level LSTM model. Our model uses a 32-dimensional word embedding table and a single layer LSTM with memory cell size 128, the exact architecture is given in Table~\ref{tab:imdb}. We experiment with the IMDB sentiment classification dataset~\citep{maas2011learning}. The dataset contains $50,000$ movie reviews collected from movie rating website IMDB. The training set has $25,000$ reviews, each labeled as positive or negative. The rest $25,000$ form the test set. We split $20\%$ of the training set to use as the development set. We refer the readers to \deepobs\citep{schneider2018deepobs} for the exact details of the other architectures used in this work.
\begin{table*}[!h]
    \caption{Architecture of the LSTM network used for IMDb experiments}
    \label{tab:imdb}
    \centering
    \begin{tabular}{l|c}
        \toprule
        \textbf{Layer name} & \textbf{Description}        \\
        \midrule
        Emb                 & $\begin{bmatrix} \ta{Embedding Layer} \\ \ta{Vocabulary of 10000}\\ \ta{Embedding dimension: 32}\end{bmatrix}$ \\ \\
        LSTM\_1             & $\begin{bmatrix} \ta{LSTM} \\ \ta{Input size: 32}\\ \ta{Hidden dimension: 128} \end{bmatrix}$ \\\\
        FC Layer            & $\lin(128\xrightarrow{}2)$  \\
        \midrule
        Classifier          & $\smax(2)$                  \\
        \bottomrule
    \end{tabular}
\end{table*}

\section{Performance Analysis}\crefalias{section}{appsec}\label{app:fullplots} 

We show the full performance plots of all variants of SGD, Adam, and Adagrad we experimented with in \cref{fig:perf_plots_app_full}. 
\begin{figure}[ht]
    \centering
    \begin{subfigure}{0.8\textwidth}
        \includegraphics[width=\textwidth ]{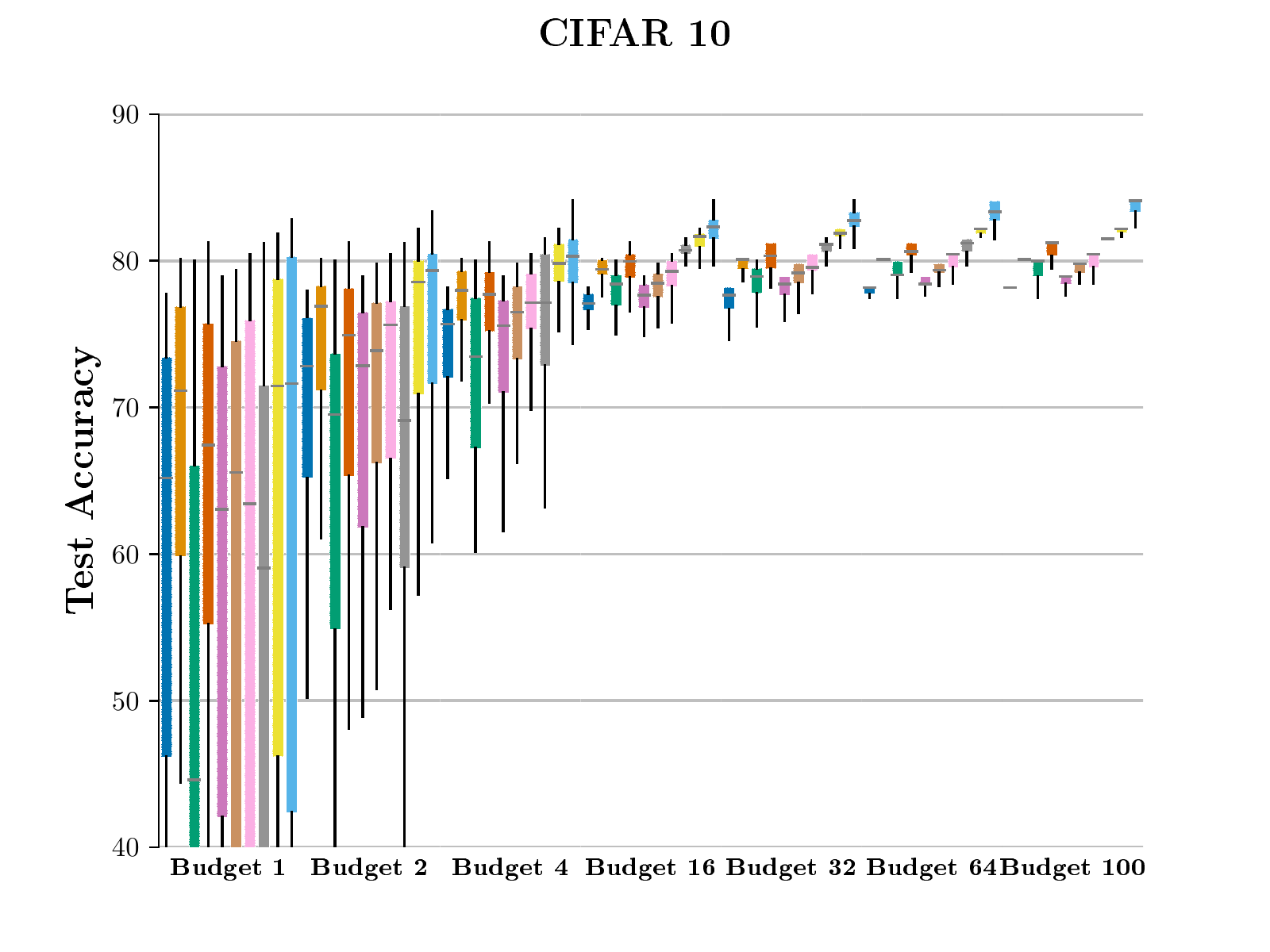}
    \end{subfigure}
    \begin{subfigure}{0.8\textwidth}
        \includegraphics[width=\textwidth ]{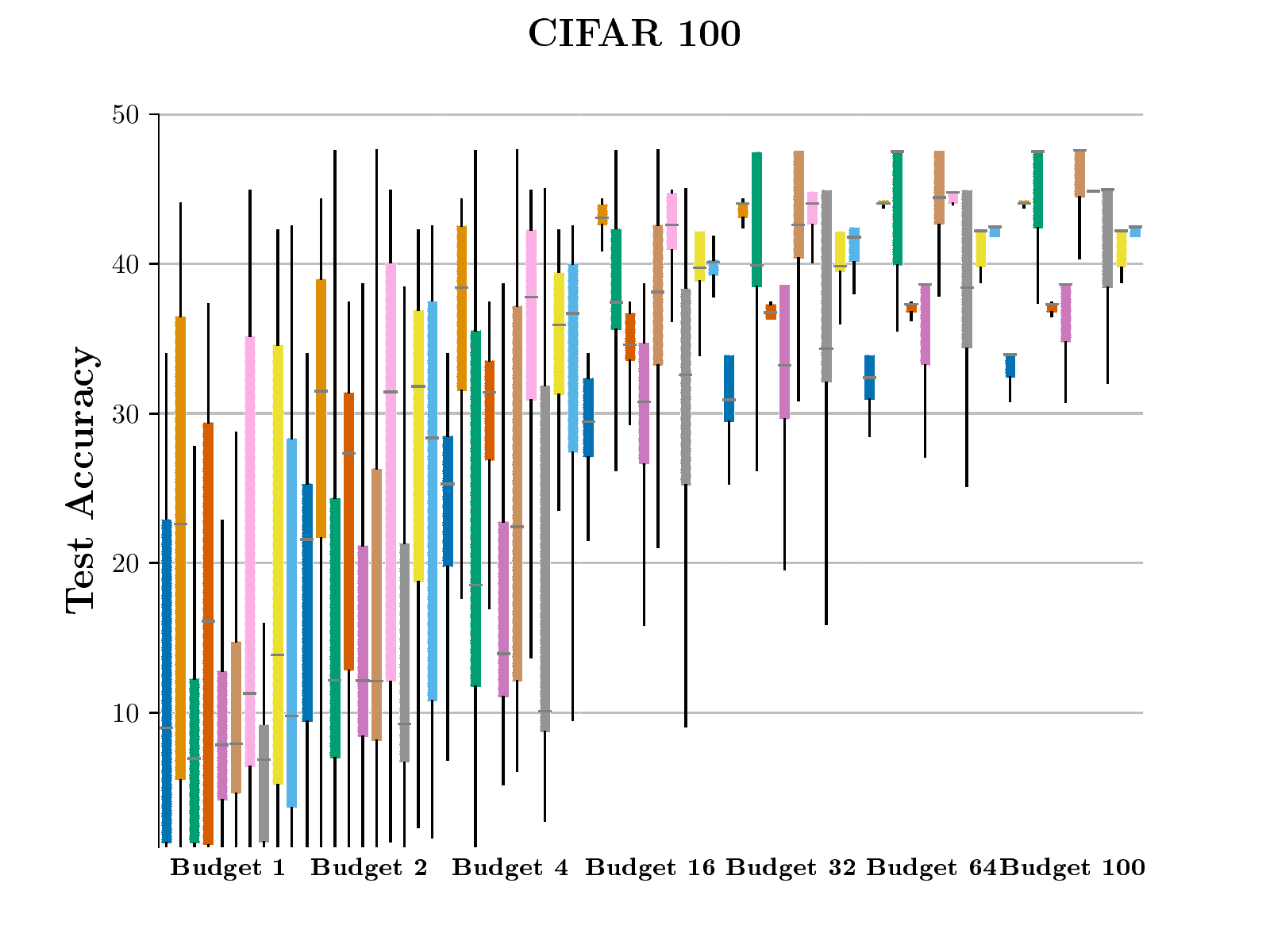}
    \end{subfigure}
    \caption{\showcolor{color1_app}{\Adagrad},
    \showcolor{color2_app}{\AdamLR},
    \showcolor{color3_app}{\Adam},
    \showcolor{color4_app}{\AdamDecay},
    \showcolor{color5_app}{\SGD},
    \showcolor{color6_app}{\SGDM},
    \showcolor{color7_app}{\SGDMC},
    \showcolor{color8_app}{\SGDMW},
    \showcolor{color9_app}{\SGDMCWC}, and
    \showcolor{color10_app}{\SGDDecay}
    }
\end{figure}
\begin{figure}[ht]\ContinuedFloat
    \centering
    \begin{subfigure}{0.8\textwidth}
        \includegraphics[width=\textwidth ]{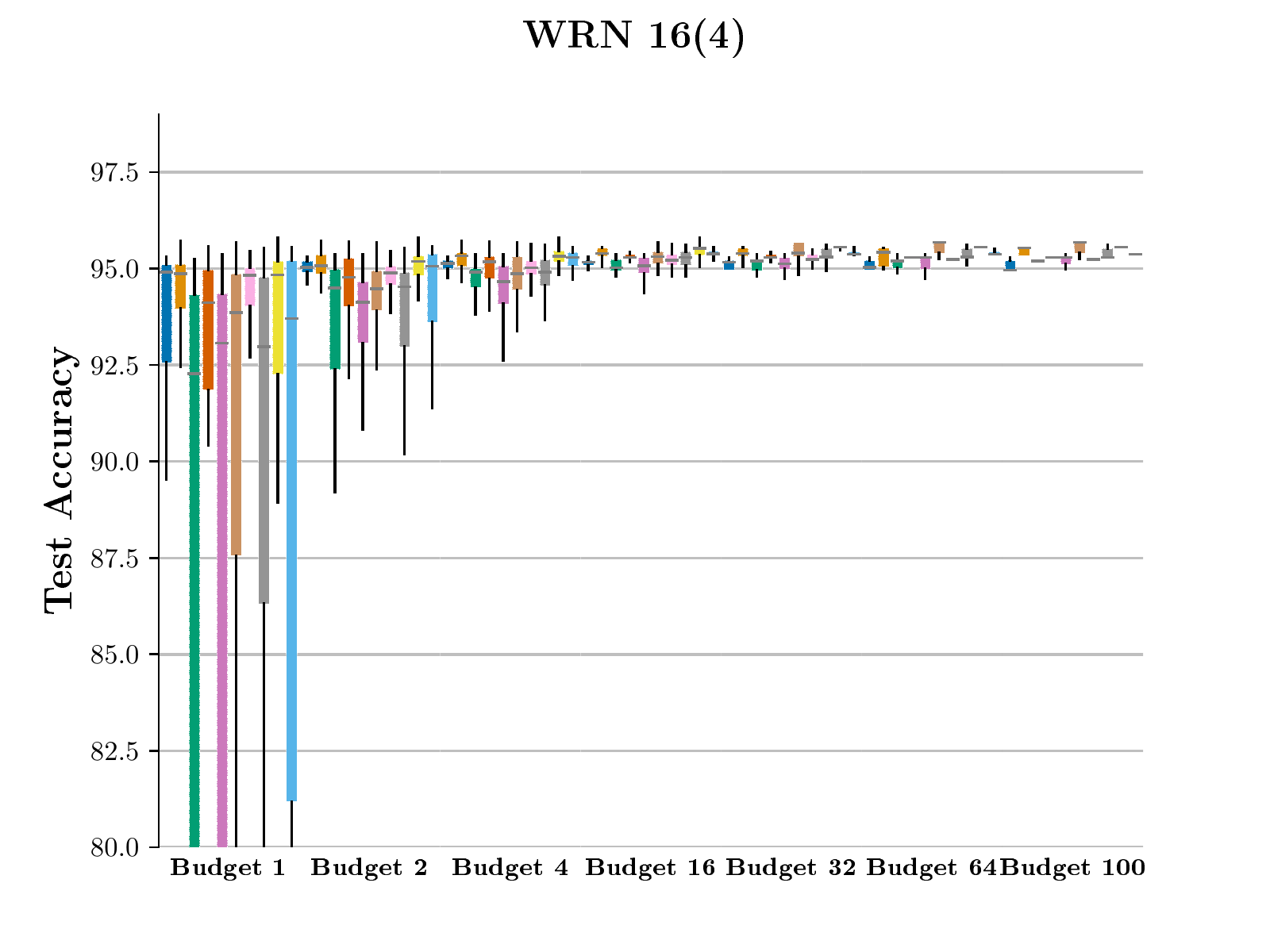}
    \end{subfigure}
    \begin{subfigure}{0.8\textwidth}
        \includegraphics[width=\textwidth ]{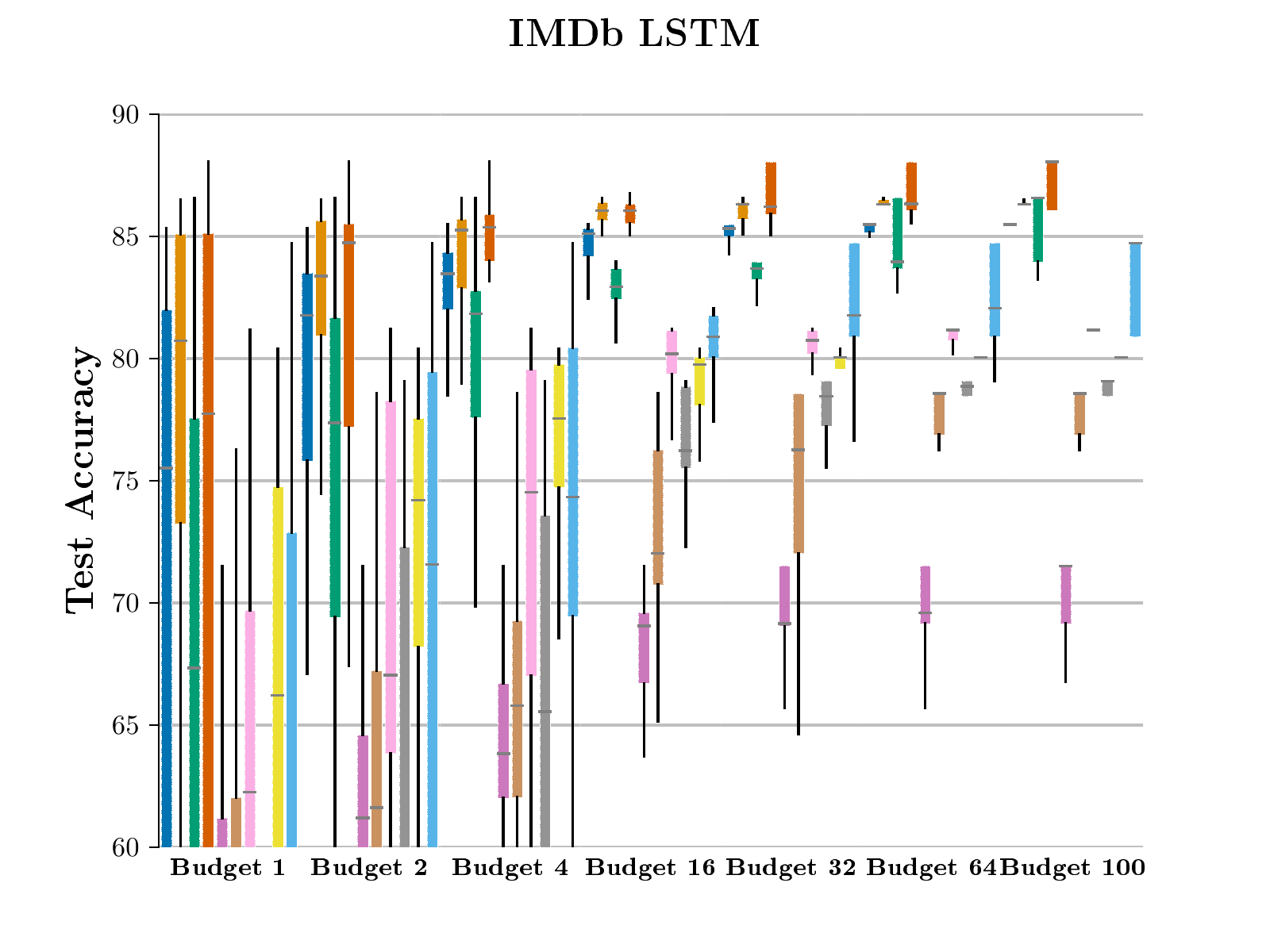}
    \end{subfigure}
    \caption{\showcolor{color1_app}{\Adagrad},
    \showcolor{color2_app}{\AdamLR},
    \showcolor{color3_app}{\Adam},
    \showcolor{color4_app}{\AdamDecay},
    \showcolor{color5_app}{\SGD},
    \showcolor{color6_app}{\SGDM},
    \showcolor{color7_app}{\SGDMC},
    \showcolor{color8_app}{\SGDMW},
    \showcolor{color9_app}{\SGDMCWC}, and
    \showcolor{color10_app}{\SGDDecay}
    }
\end{figure}
\begin{figure}[ht]\ContinuedFloat
    \centering
    \begin{subfigure}{0.8\textwidth}
        \includegraphics[width=\textwidth ]{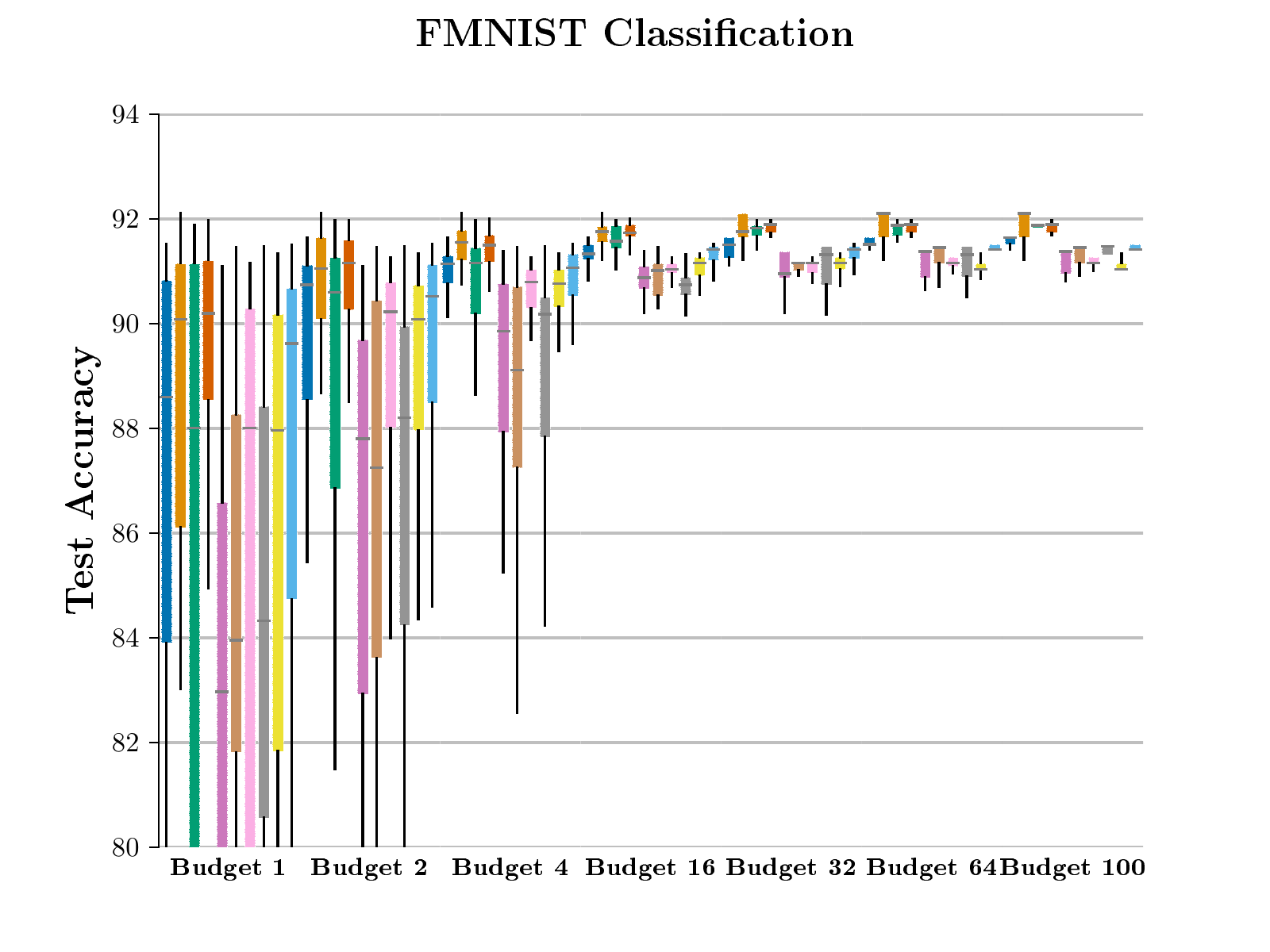}
    \end{subfigure}
    \begin{subfigure}{0.8\textwidth}
        \includegraphics[width=\textwidth ]{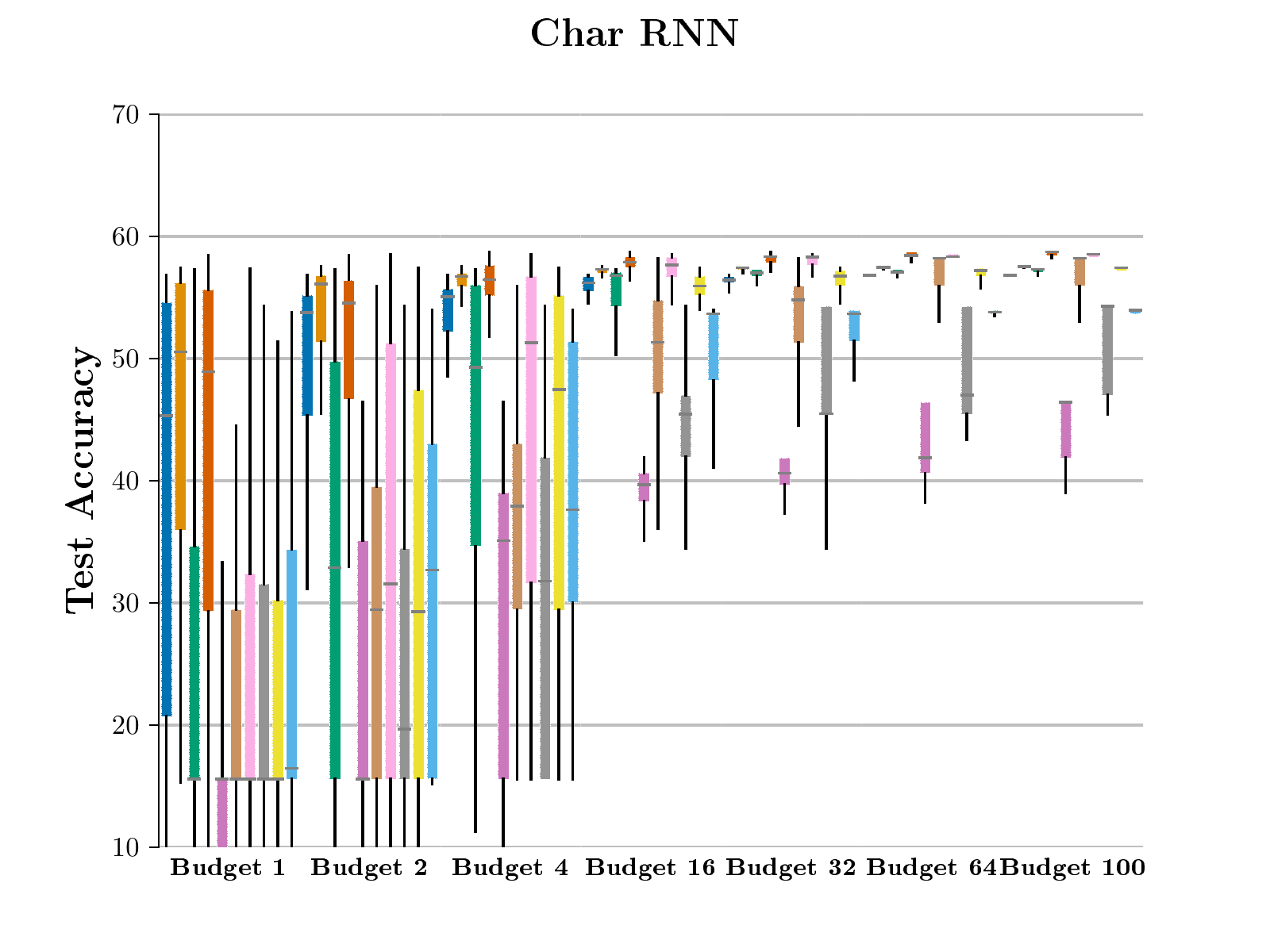}
    \end{subfigure}
    \caption{\showcolor{color1_app}{\Adagrad},
    \showcolor{color2_app}{\AdamLR},
    \showcolor{color3_app}{\Adam},
    \showcolor{color4_app}{\AdamDecay},
    \showcolor{color5_app}{\SGD},
    \showcolor{color6_app}{\SGDM},
    \showcolor{color7_app}{\SGDMC},
    \showcolor{color8_app}{\SGDMW},
    \showcolor{color9_app}{\SGDMCWC}, and
    \showcolor{color10_app}{\SGDDecay}
    }
\end{figure}
\begin{figure}[ht]\ContinuedFloat
    \centering
    \begin{subfigure}{0.8\textwidth}
        \includegraphics[width=\textwidth ]{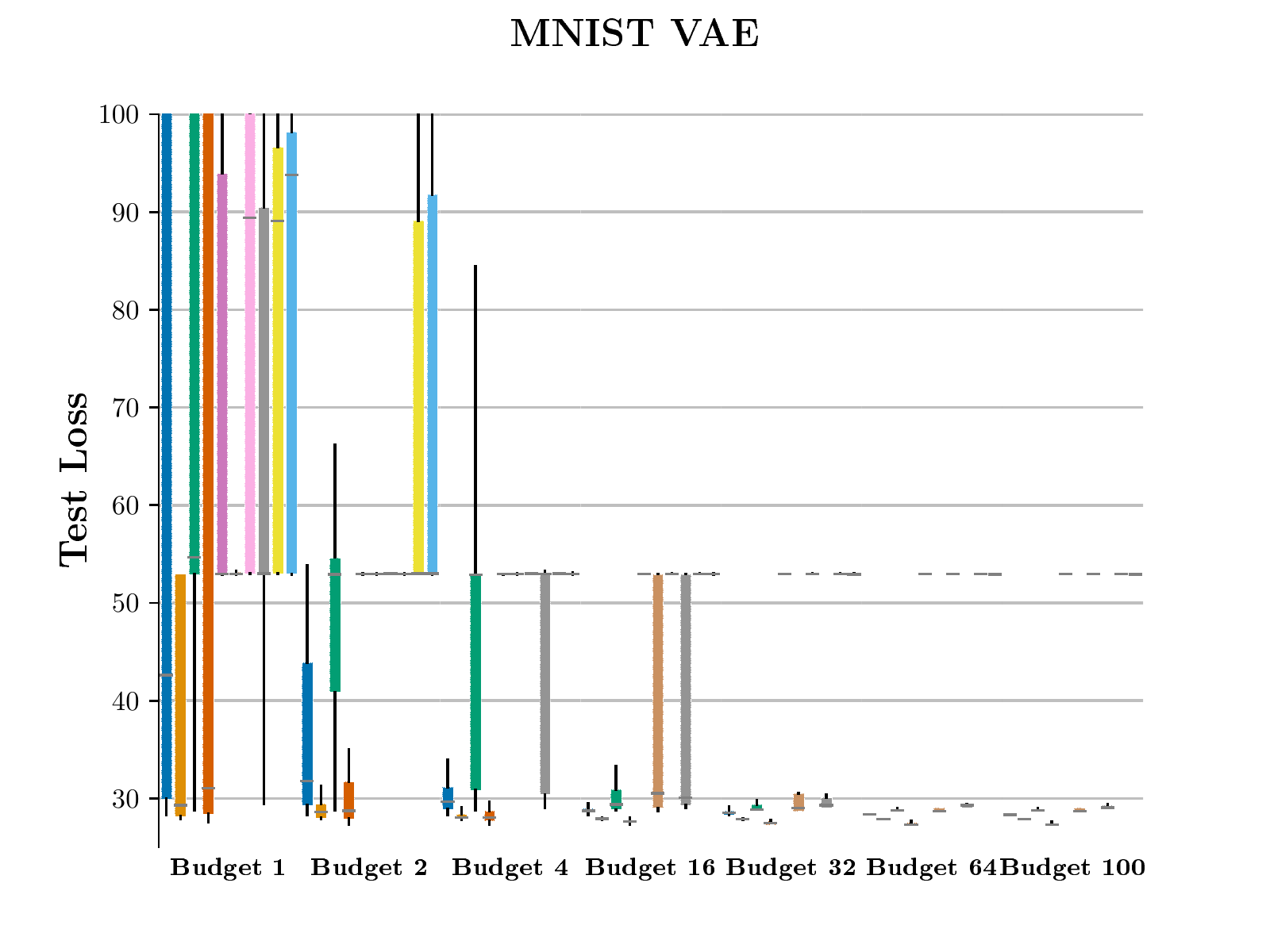}
    \end{subfigure}
    \begin{subfigure}{0.8\textwidth}
        \includegraphics[width=\textwidth ]{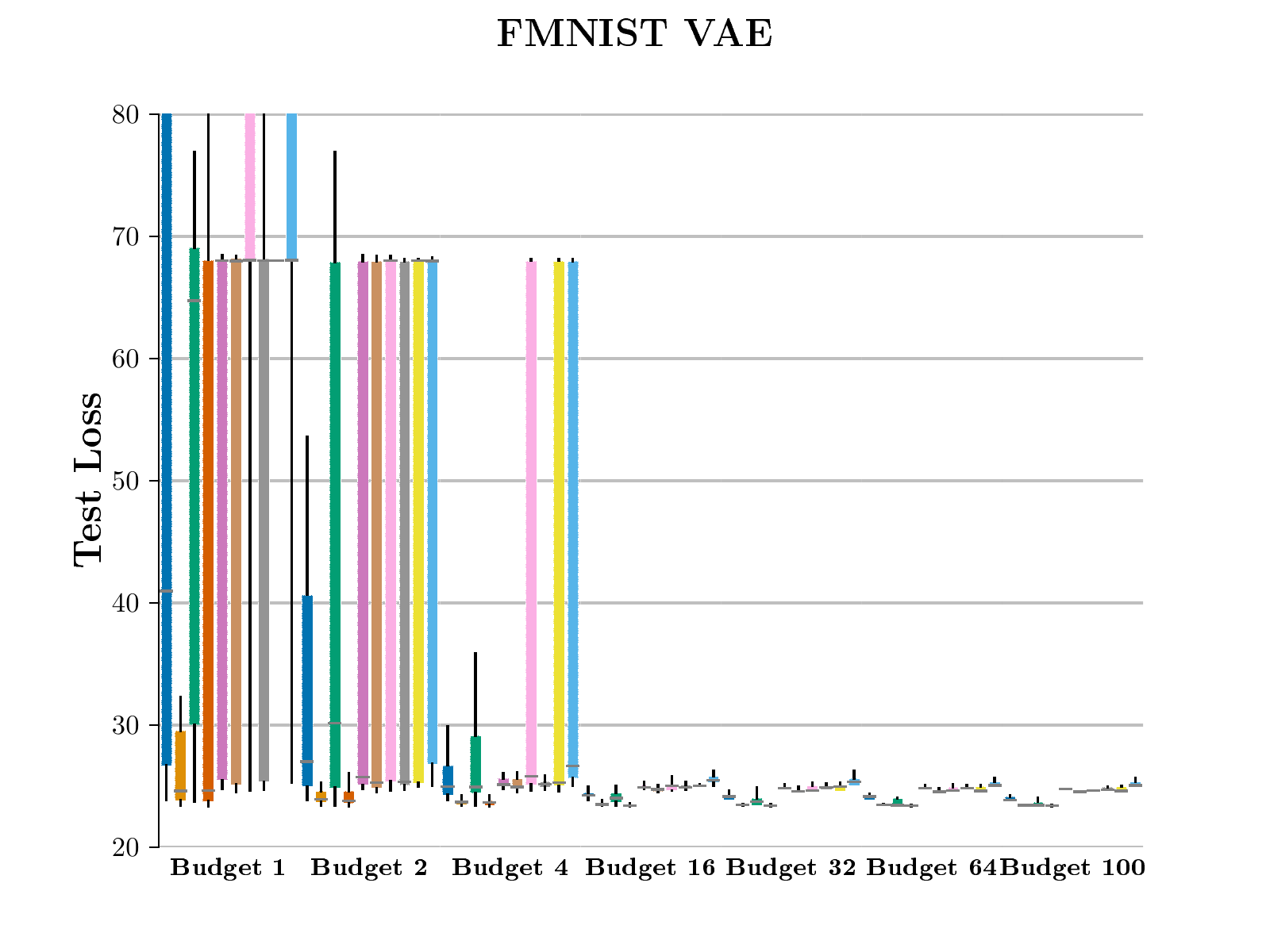}
    \end{subfigure}
    \caption{\showcolor{color1_app}{\Adagrad},
    \showcolor{color2_app}{\AdamLR},
    \showcolor{color3_app}{\Adam},
    \showcolor{color4_app}{\AdamDecay},
    \showcolor{color5_app}{\SGD},
    \showcolor{color6_app}{\SGDM},
    \showcolor{color7_app}{\SGDMC},
    \showcolor{color8_app}{\SGDMW},
    \showcolor{color9_app}{\SGDMCWC}, and
    \showcolor{color10_app}{\SGDDecay}
    }
\end{figure}
\begin{figure}[ht]\ContinuedFloat
    \centering
    \begin{subfigure}{0.8\textwidth}
        \includegraphics[width=\textwidth ]{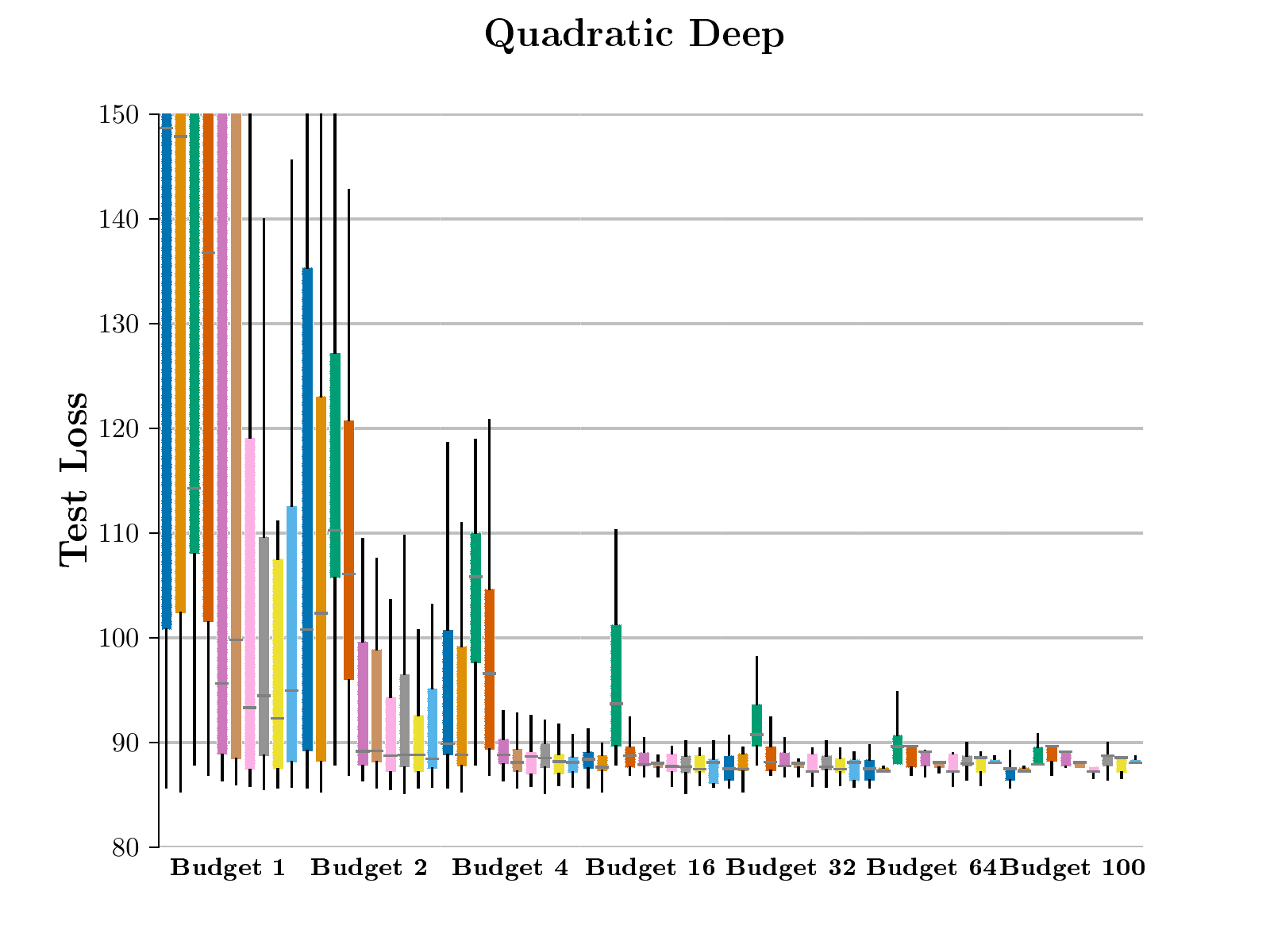}
    \end{subfigure}
    \caption{We show the performance of \showcolor{color1_app}{\Adagrad},
    \showcolor{color2_app}{\AdamLR},
    \showcolor{color3_app}{\Adam},
    \showcolor{color4_app}{\AdamDecay},
    \showcolor{color5_app}{\SGD},
    \showcolor{color6_app}{\SGDM},
    \showcolor{color7_app}{\SGDMC},
    \showcolor{color8_app}{\SGDMW},
    \showcolor{color9_app}{\SGDMCWC}, and
    \showcolor{color10_app}{\SGDDecay}
    over all the experiments. We plot the on the x-axis the number of the hyperparameter configuration searches, on the y-axis the appropriate performance.}
    \label{fig:perf_plots_app_full}
\end{figure}

\section{How Likely Are We to Find Good Configurations?}   \crefalias{section}{appsec}

\label{app:probab_best_config}
In \cref{fig:prob_plot} we showed the chance of finding the optimal hyperparameter setting for some of the optimizers considered, in a problem agnostic setting. Here we delve into the case where we present similar plots for each of the problems considered in \cref{sec:experiments}.

\begin{figure*}[ht]
    \centering
    \includegraphics[width=\textwidth]{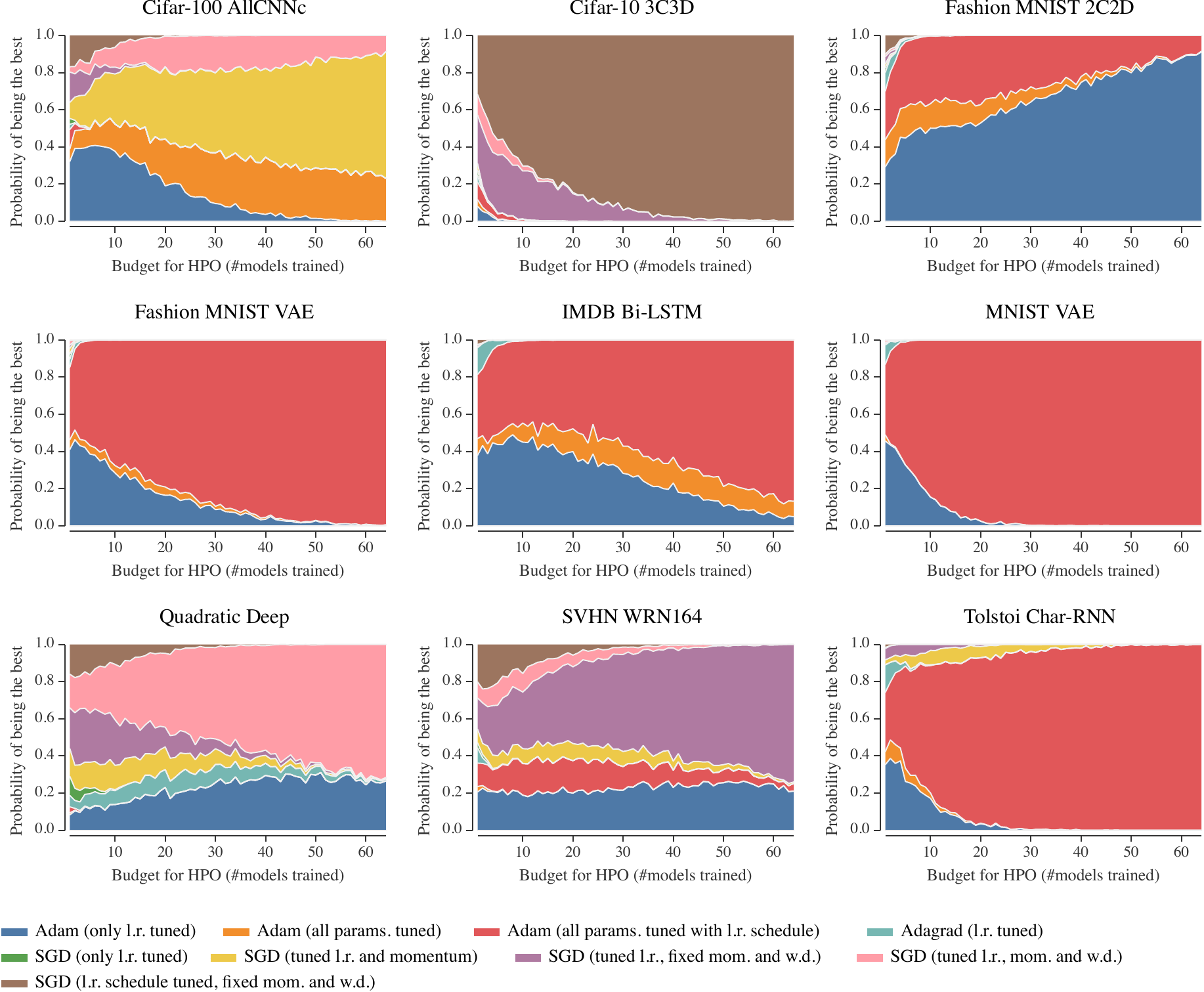}
    \caption{
        Which optimizer for which budget? Given a tuning budget $K$ ($x$-axis), the stacked area plots above show how likely each optimizer (colored bands) is to yield the best result after $K$ steps of hyperparameter optimization. 
        For example, for the IMDB LSTM problem, for a small budget, \AdamLR is the best choice (with $\sim 0.8$ probability), whereas for a larger search budget $> 50$, tuning the other parameters of `Adam' is likely to pay off.
    }
    \label{fig:my_label}
\end{figure*}

A natural question that arises is: Given a budget $K$, what is the best optimizer one can pick? In other words, for a given budget what is the probability of each optimizer finding the best configuration? We answer this with a simple procedure. We repeat the runs of HPO for a budget $K$, and collect the optimizer that gave the best result in each of those runs. Using the classical definition of probability, we compute the required quantity. We plot the computed probability in Figure~\ref{fig:my_label}. It is very evident for nearly all budgets, \AdamLR is always the best option for 4 of the problems. SGD variants emerge to be better options for CIFAR-100 and Char-RNN at later stages of HPO. For some of the problems like VAEs, LSTM, it is very obvious that \AdamLR is nearly always the best choice. This further strengthens our hypothesis that adaptive gradient methods are more tunable, especially in constrained HPO budget scenarios.

\section{Computing the Expected Maximum of Random Samples}   \crefalias{section}{appsec} \label{app:expmax}

The following is a constructive proof of how to compute the expected value that the bootstrap method converges to in the limit of infinite re-sampling. It is a paraphrase of \citet[Section~3.1]{dodge2019show}, but due to inaccuracies in \cref{eq:emp_distrib} in their paper, we repeat it here for clarity. 

Let $x_1, x_2 \dots x_N \sim \mathcal{X}$ be $N$ independently sampled values. Let the random variable $\mathbf{Y}$ be the maximum of a random subset of size $S$ from $x_1, x_2 \dots x_N$ where $S \leq N$. For representational convenience, let them be the first $S$ samples. So, $\mathbf{Y} = \max \{x_1, \dots, x_S\}$. We are interested in computing $\mathbb{E}[\mathbf{Y}]$. This can be computed as \[\mathbb{E}[\mathbf{Y}] = \sum_y y \cdot P(\mathbf{Y} = y)\] for discrete $\mathbf{Y}$, with $P(\mathbf{Y} = y)$ be the probability mass function of $\mathbf{Y}$. We can write \[P(\mathbf{Y} = y) = P(\mathbf{Y} \leq y) - P(\mathbf{Y} < y)\]

As $x_i ~\forall i$ are iid sampled, 
\begin{align*}
    P(\mathbf{Y} \leq y) &= P(\max_{i=1\dots S} x_i \leq y) \\
    &= \prod_{i=1}^S P(x_i \leq y) \\
    &= P(X \leq y)^S
\end{align*}

$P(X \leq y)$ can be empirically estimated from data as the sum of normalized impulses. 
\begin{equation} \label{eq:emp_distrib}
   P(X\leq y) = \frac{1}{N}\sum_{i=1}^N \mathbb{I}_{x_i\leq y}  
\end{equation}

Thus,
\begin{equation} \label{eq:avgofmax}
    \mathbb{E}[\mathbf{Y}] =  \sum_y y (P(X\leq y)^S - P(X<y)^S )    
\end{equation}

A very similar equation can be derived to compute the variance too. Variance is defined as $Var(\mathbf{Y}) = \mathbb{E}[Y^2] - \mathbb{E}[Y]^2$. The second operand is given by \cref{eq:avgofmax}. The first operand (for discrete distributions) can be computed as

\begin{equation} \label{eq:varofmax}
    \mathbb{E}[\mathbf{Y}^2] =  \sum_y y^2 (P(X\leq y)^S - P(X<y)^S )      
\end{equation}

Given the iterates (not incumbents) of Random Search, the expected performance at a given budget can be estimated by \cref{eq:avgofmax} and the variance can be computed by \cref{eq:varofmax}.

\section{Aggregating the Performance of Incumbents}\crefalias{section}{appsec}
\label{app:aggregation}
In Procedure~\ref{alg:benchmark}, we propose returning all the incumbents of the HPO algorithm. Here we propose the use of an aggregation function that helps create a comparable scalar that can used in a benchmarking software like \deepobs to rank the performance of optimizers that takes into cognizance the ease-of-use aspect too. 

\subsection{Aggregating Function}\label{sec:agg_fun}
For the aggregation function discussed, we propose a simple convex combination of the incumbent performances and term it $\omega$-tunability. If $\mathcal{L}_t$ be the incumbent performance at budget $t$, we define $\omega$-tunability as \[\omega\text{-tunability} = \sum_{t=1}^T \omega_t \mathcal{L}_t\] where $w_t >0$ $\forall$ t and $\sum_t w_t =1$

By appropriately choosing the weights $\{\omega_t\}$, we can interpolate between our two notions of tunability in \cref{sec:whyhpo}. In the extreme case where we are only interested in the peak performance of the optimizer, we can set $\omega_T = 1$ and set the other weights to zero. In the opposite extreme case where we are interested in the "one-shot tunability" of the optimizer, we can set $\omega_1 = 1$. In general, we can answer the question of "How well does the optimizer perform with a budget of $K$ runs?" by setting \impulseatk. \cref{fig:perf_plots} and \cref{fig:perf_plots_app_full} can also be computed as \impulseatk.

While the above weighting scheme is intuitive, merely computing the performance after expending HPO budget of $K$ does not consider the performance obtained after the previous $K-1$ iterations i.e. we would like to differentiate the cases where a requisite performance is attained by tuning an optimizer for $K$ iterations and another for $K_1$ iterations, where $K_1\gg K$. Therefore, we employ a weighting scheme as follows: By setting $\omega_i \propto (T - i)$, our first one puts more emphasis on the earlier stages of the hyperparameter tuning process. We term this weighting scheme \emph{Cumulative Performance-Early}(\weightingearly).  %
The results of the various optimzers' \weightingearly~ is shown in \cref{tab:cpe}. It is very evident that \AdamLR fares the best across tasks. Even when it is not the best performing one, it is quite competitive. Thus, our observation that \AdamLR is the easiest-to-tune i.e. it doesn't guarantee the best performance, but it gives very competitive performances early in the HPO search phase, holds true. 

For a benchmarking software package like \deepobs, we suggest the use of \weightingearly ~to rank optimziers, as it places focus on ease-of-tuning. This supplements the existing peak performance metric reported previously.

\begin{table*}[ht]
    \centering
    \resizebox{\textwidth}{!}{
        \begin{tabular}{@{}lllllll|lll@{}}
            \toprule
            \textbf{Optimizer} & \textbf{FMNIST(\%)$\uparrow$} & \textbf{CIFAR 10(\%)$\uparrow$} & \textbf{CIFAR 100(\%)$\uparrow$} & \textbf{IMDb(\%)$\uparrow$} & \textbf{WRN 16(4)(\%)$\uparrow$} & \textbf{Char-RNN(\%)$\uparrow$} & \textbf{MNIST-VAE$\downarrow$} & \textbf{FMNIST-VAE$\downarrow$} & \textbf{Quadratic Deep$\downarrow$} \\ \midrule
            \AdamLR            & \textbf{91.6}   & 78.8              & \textbf{42.0}      & 85.9          & \textbf{95.3}      & 56.9              & 28.9               & \textbf{24.3}       & 89.9                    \\
            \Adam              & 91.3            & 77.3              & 38.1               & 83.4          & 94.5               & 54.2              & 33.1               & 25.7                & 95.4                    \\
            \SGDMCWC           & 90.8            & 81.0              & 38.8               & 78.7          & \textbf{95.3}      & 53.9              & 54.0               & 27.9                & \textbf{87.4}           \\
            \SGDMW             & 90.5            & 79.6              & 33.2               & 75.2          & 95.0               & 44.4              & 35.2               & 26.5                & 87.5                    \\
            \SGDDecay          & 91.1            & \textbf{82.1}     & 39.2               & 80.5          & 95.2               & 49.6              & 54.3               & 29.8                & 87.5                    \\ \midrule
            \Adagrad           & 91.3            & 76.6              & 29.8               & 84.4          & 95.0               & 55.6              & 30.7               & 25.9                & 90.6                    \\
            \AdamDecay         & \textbf{91.6}   & 79.4              & 35.1               & \textbf{86.0} & 95.1               & \textbf{57.4}     & \textbf{28.6}      & \textbf{24.3}       & 92.8                    \\
            \SGD               & 90.4            & 76.9              & 30.6               & 68.1          & 94.7               & 39.9              & 53.4               & 26.2                & 89.3                    \\
            \SGDM              & 90.5            & 77.8              & 39.8               & 73.8          & 94.9               & 50.7              & 37.1               & 26.3                & 88.2                    \\
            \SGDMC             & 90.7            & 78.8              & \textbf{42.0}      & 79.0          & 95.0               & 55.5              & 54.1               & 28.5                & 88.1                    \\ \bottomrule
            \end{tabular}}
    \caption{\weightingearly ~for the various optimizers experimented. It is evident that \AdamLR is the most competitive across tasks.}
    \label{tab:cpe}
\end{table*}

\section{Results for Computation Time Budgets}\label{sec:time-budget}
Using number of hyperparameter configuration trials as budget unit does not account for the possibility that optimizers may require different amounts of computation time to finish a trial. 
While the cost for one update step is approximately the same for all optimizers, some require more update steps than others before reaching convergence.

To verify that our results and conclusions are not affected by our choice of budget unit, we simulate the results we would have obtained with a computation time budget in the following way.
For a given test problem (e.g., CIFAR-10), we compute the minimum number of update steps any optimizer has required to finish 100 trials, and consider this number to be the maximum computation budget.
We split this budget into 100 intervals of equal size.
Using the bootstrap~\cite{tibshirani1993introduction}, we then simulate 1,000 HPO runs, and save the best performance achieved at each interval. 
Note that sometimes an optimizer can complete multiple hyperparameter trials in one interval, and sometimes a single trial may take longer than one interval.
Finally, we average the results from all 1,000 HPO runs and compute the same summary across datasets as in Section~\ref{sec:summary_stats}.

\begin{figure}[htpb]
  \begin{center}
    \includegraphics[width=0.40\textwidth]{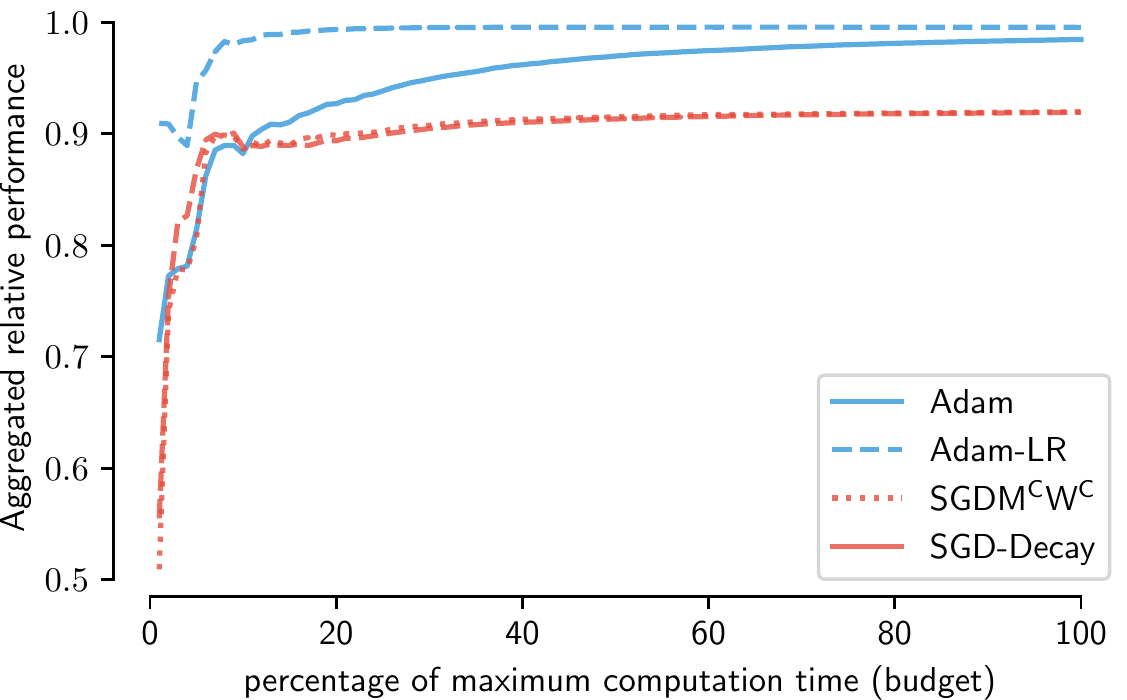}
  \end{center}
  \vspace{-0.5\baselineskip}
  \caption{Aggregated relative performance of each optimizer across datasets.}
  \label{fig:sumstat_epochs}
\end{figure}

Figure~\ref{fig:sumstat_epochs} shows that the conclusions do not change when using computation time as budget unit. In fact, the graphs show almost the exact same pattern as in Figure~\ref{fig:sumstat_}, where number of hyperparameter trials is the budget unit.

\section{Plotting Hyperparameter Surfaces}\label{sec:app-surfaces}
In Section~\ref{sec:whyhpo}, we hypothesize that the performance as a function of the hyperparameter, e.g., learning rate, of an optimizer that performs well with few trials has a wider extremum compared to an optimizer that only performs well with more trials.

\begin{figure*}[htp]
    \begin{subfigure}[t]{0.33\textwidth}
        \includegraphics[width=\textwidth]{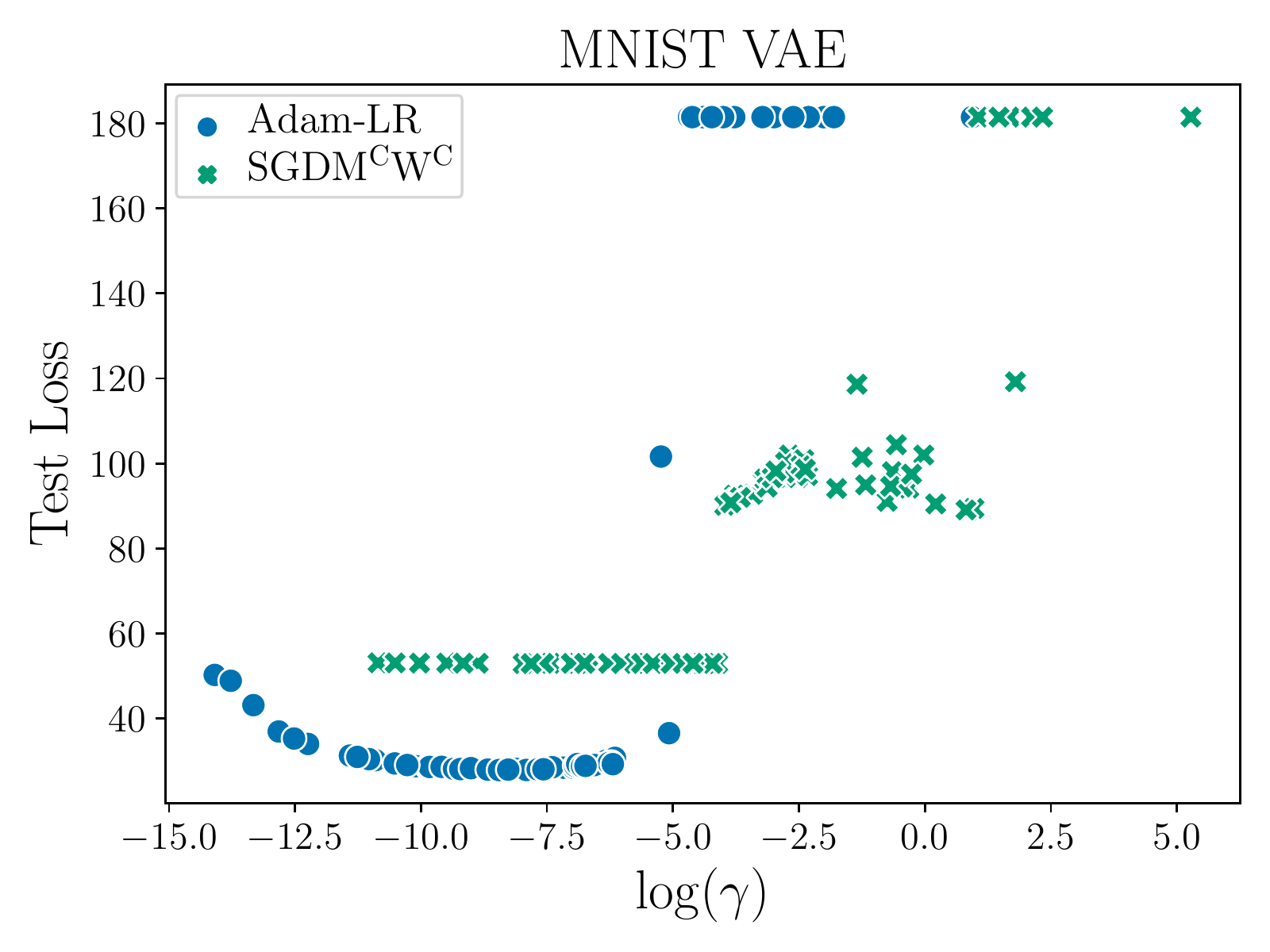}
    \end{subfigure}
    \begin{subfigure}[t]{0.33\textwidth}
        \includegraphics[width=\textwidth]{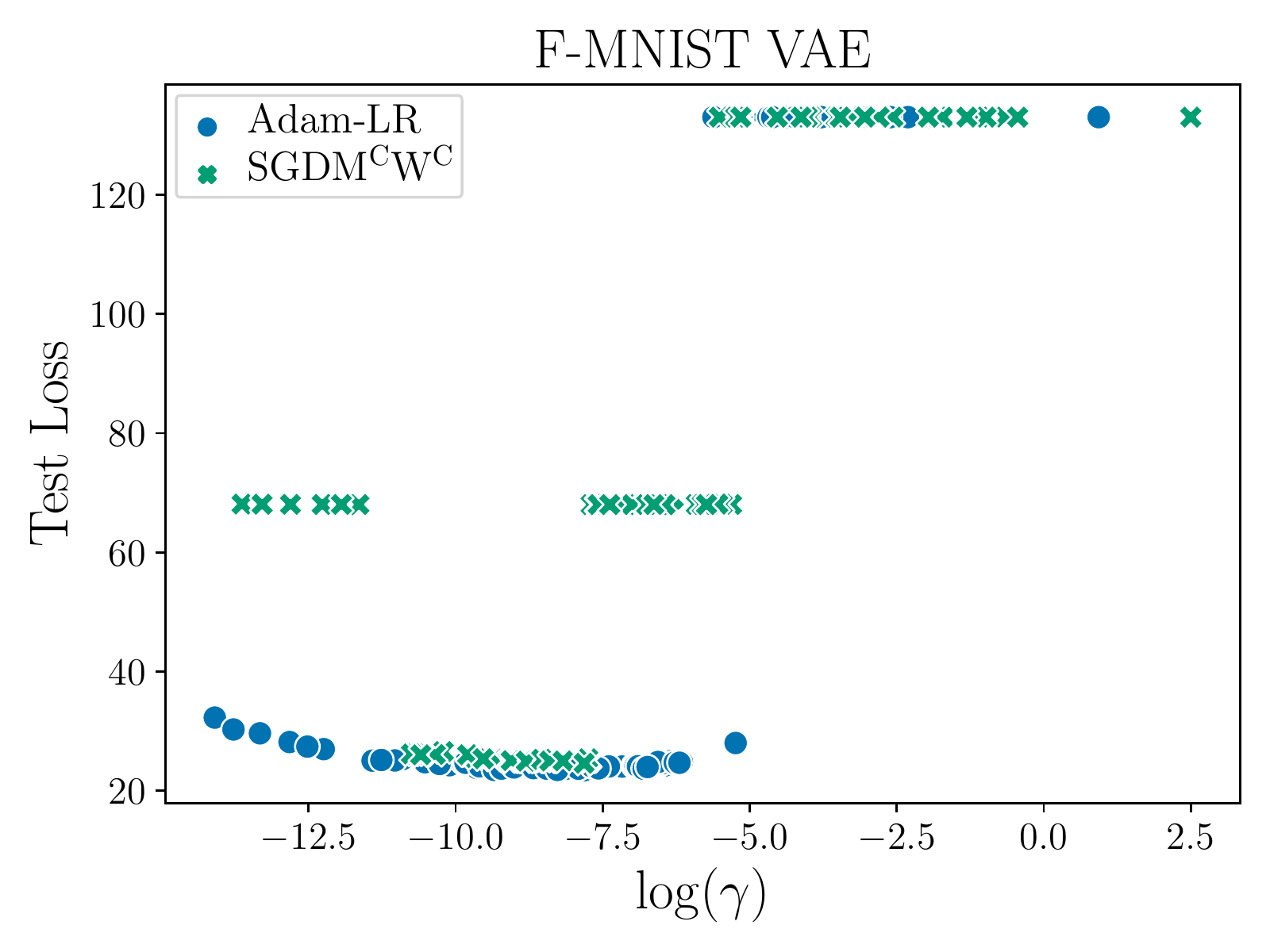}
    \end{subfigure}
    \begin{subfigure}[t]{0.33\textwidth}
        \includegraphics[width=\textwidth]{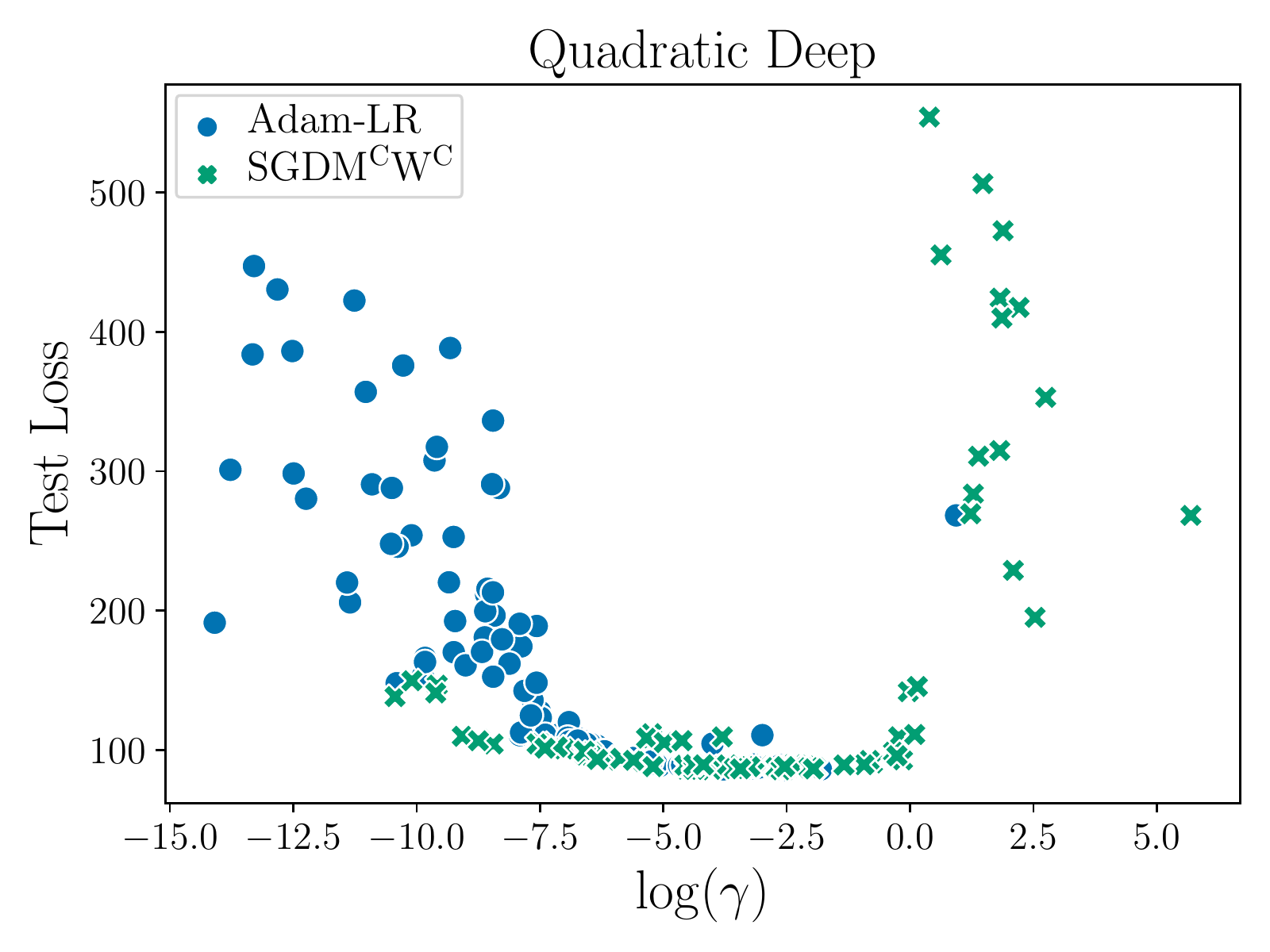}
    \end{subfigure}
    \begin{subfigure}[t]{0.33\textwidth}
        \includegraphics[width=\textwidth]{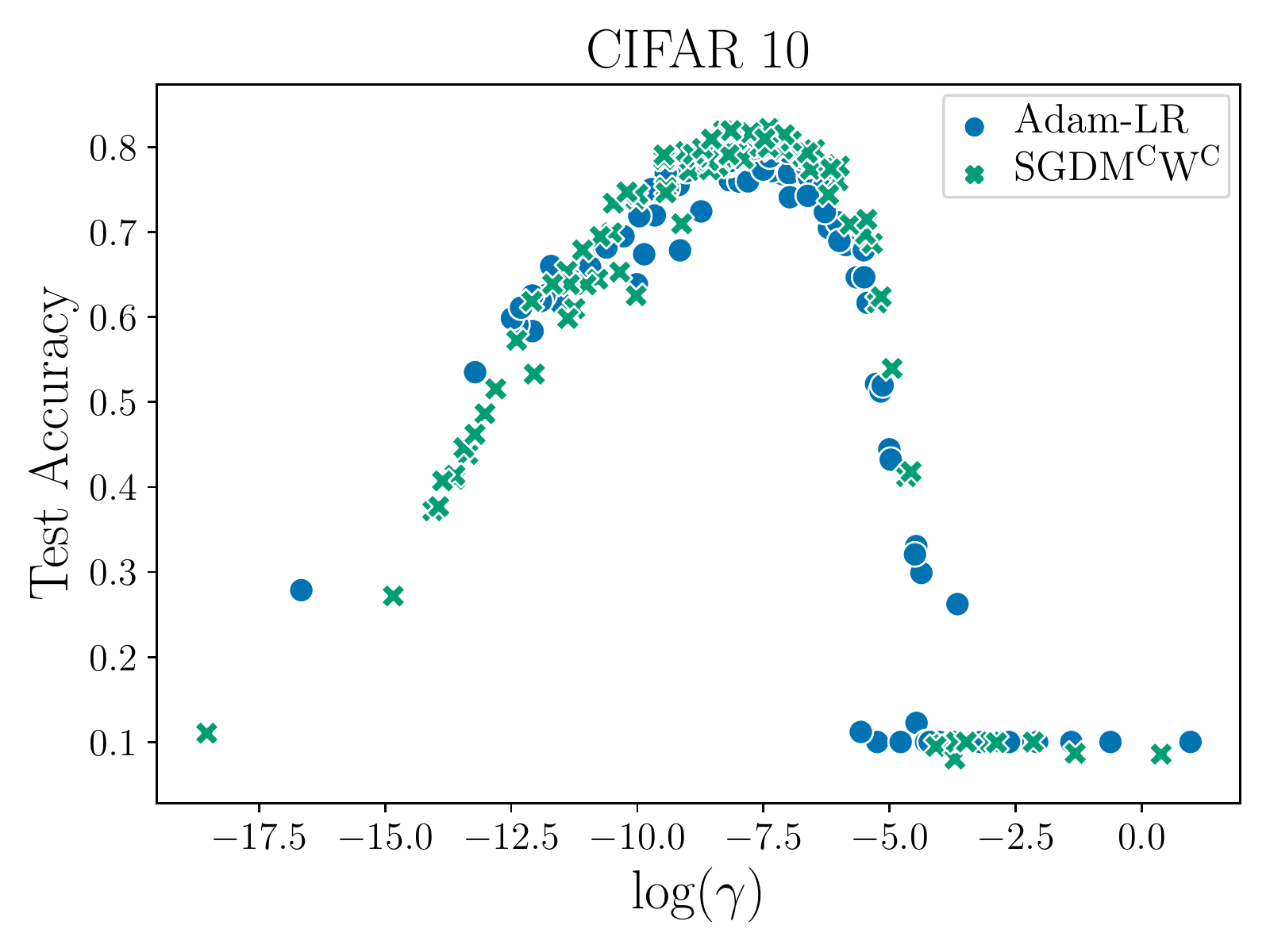}
    \end{subfigure}
    \begin{subfigure}[t]{0.33\textwidth}
        \includegraphics[width=\textwidth]{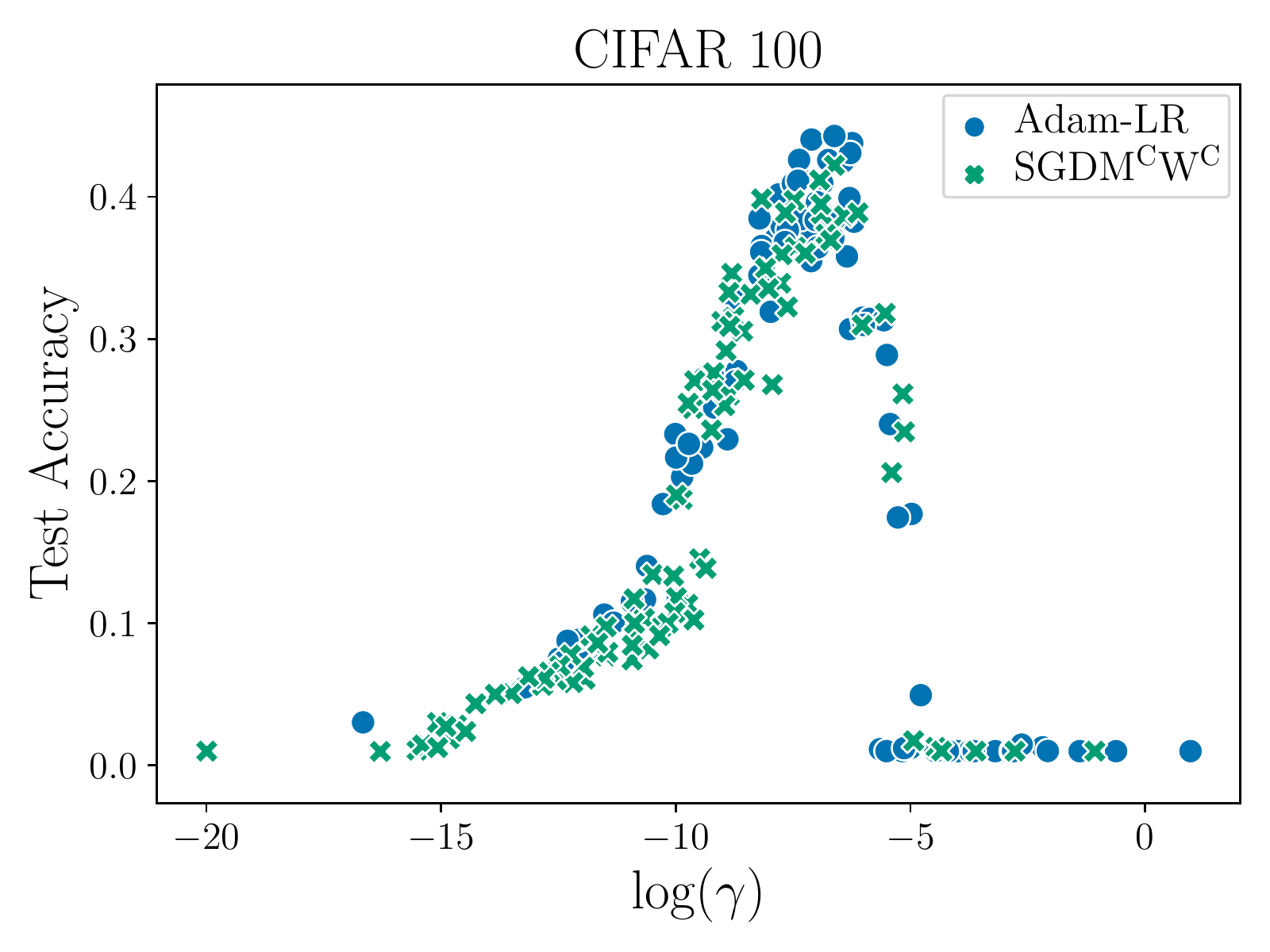}
    \end{subfigure}
    \begin{subfigure}[t]{0.33\textwidth}
        \includegraphics[width=\textwidth]{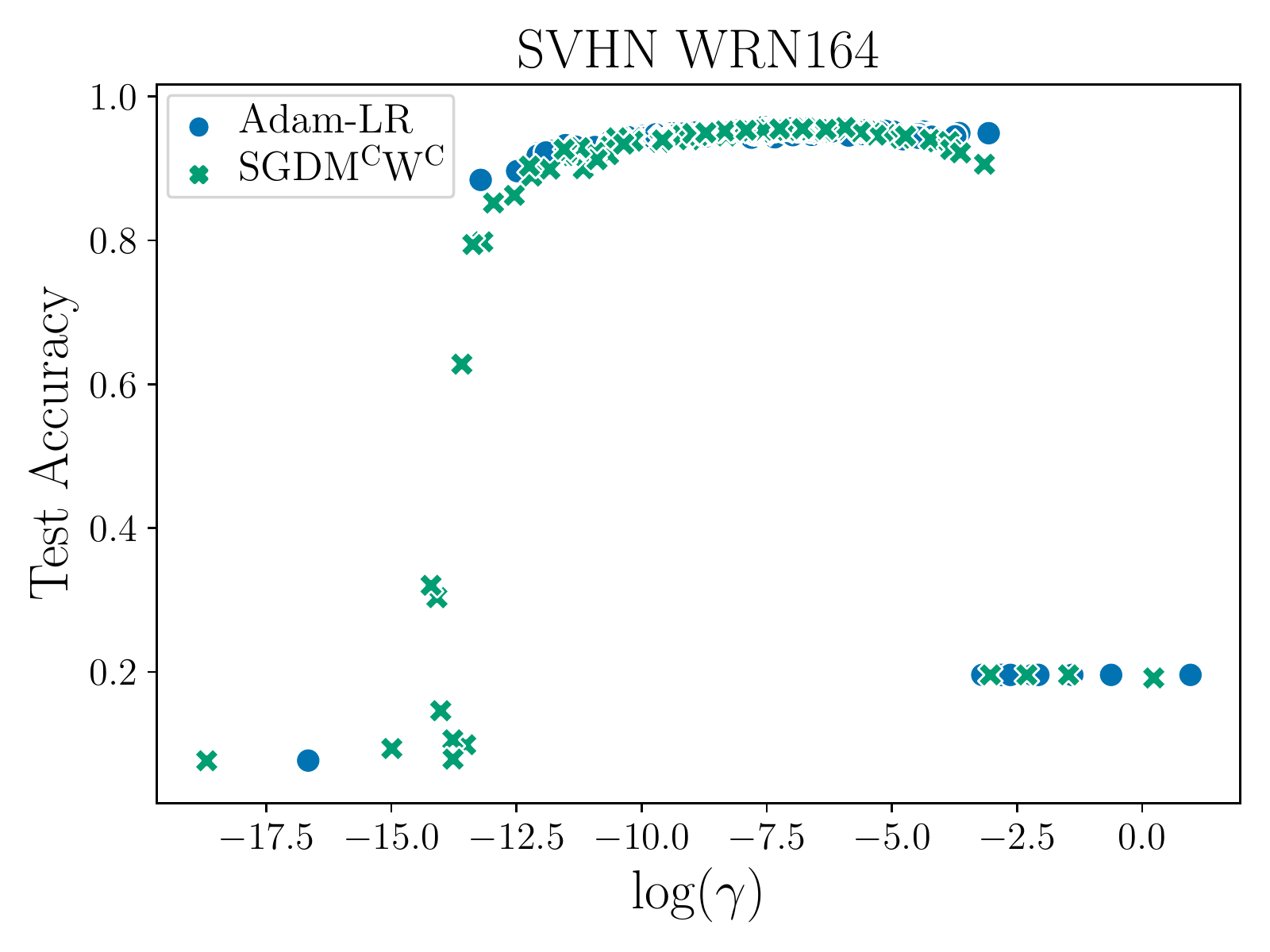}
    \end{subfigure}
    \begin{subfigure}[t]{0.33\textwidth}
        \includegraphics[width=\textwidth]{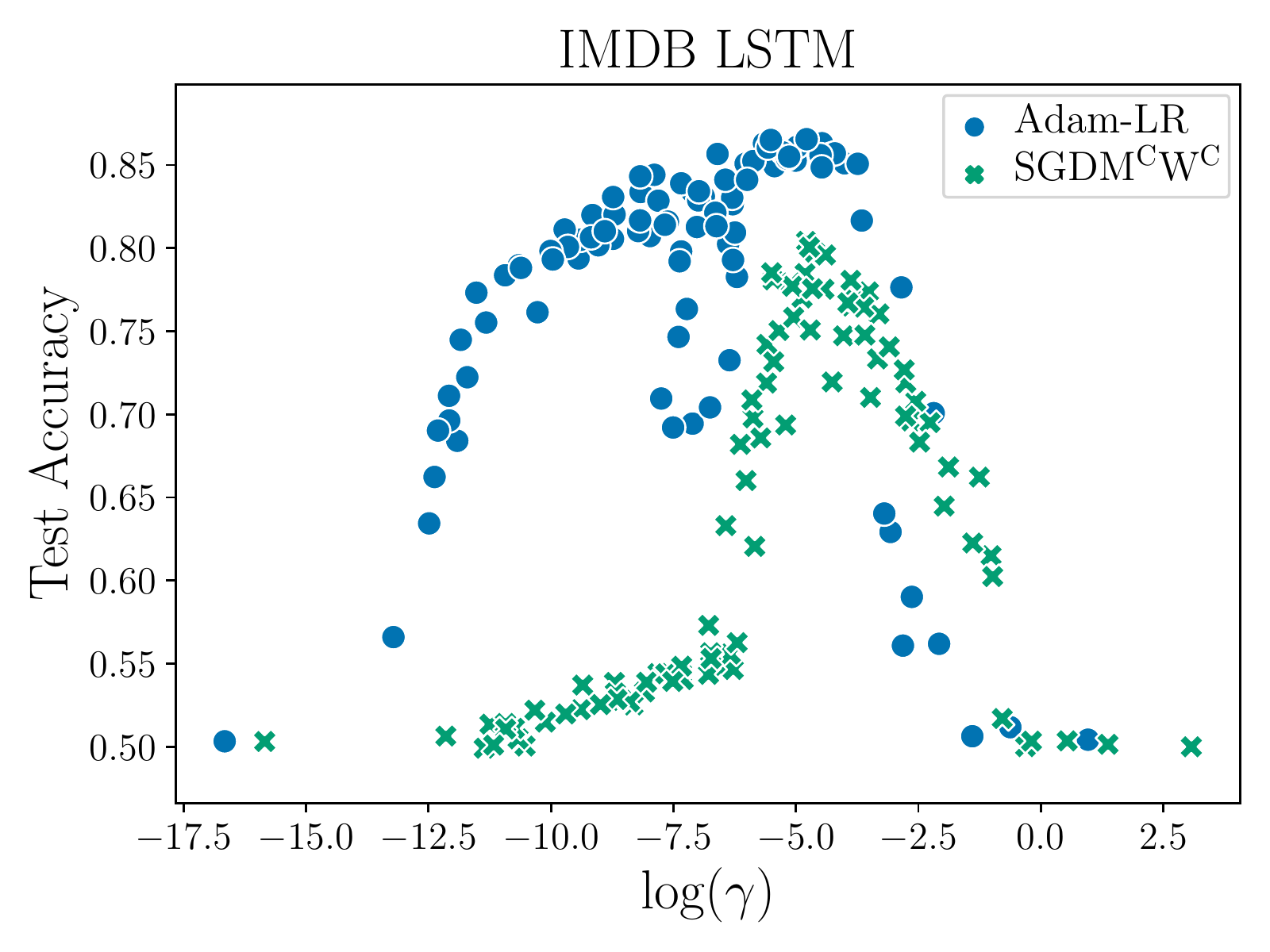}
    \end{subfigure}
    \begin{subfigure}[t]{0.33\textwidth}
        \includegraphics[width=\textwidth]{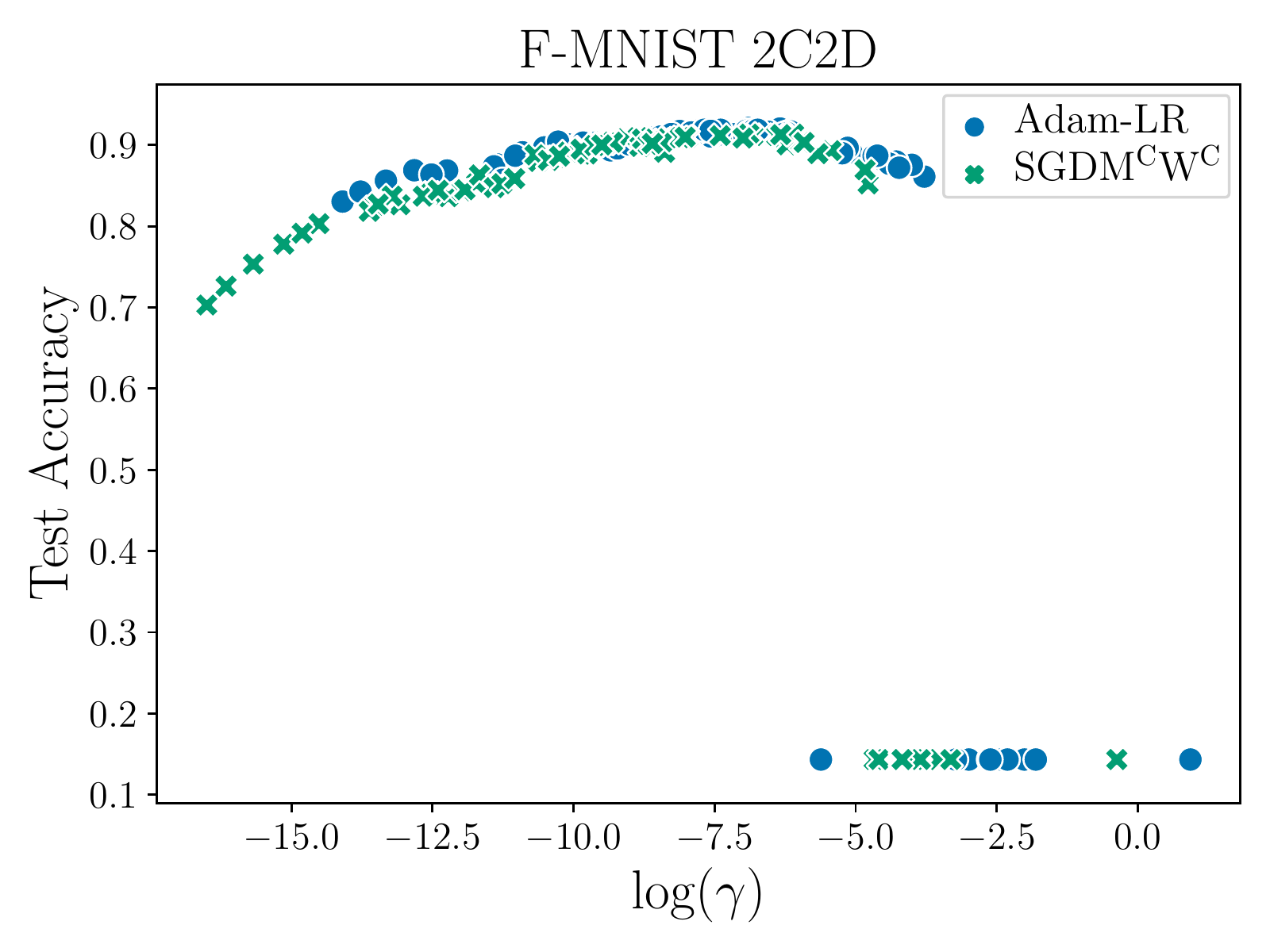}
    \end{subfigure}
    \begin{subfigure}[t]{0.33\textwidth}
        \includegraphics[width=\textwidth]{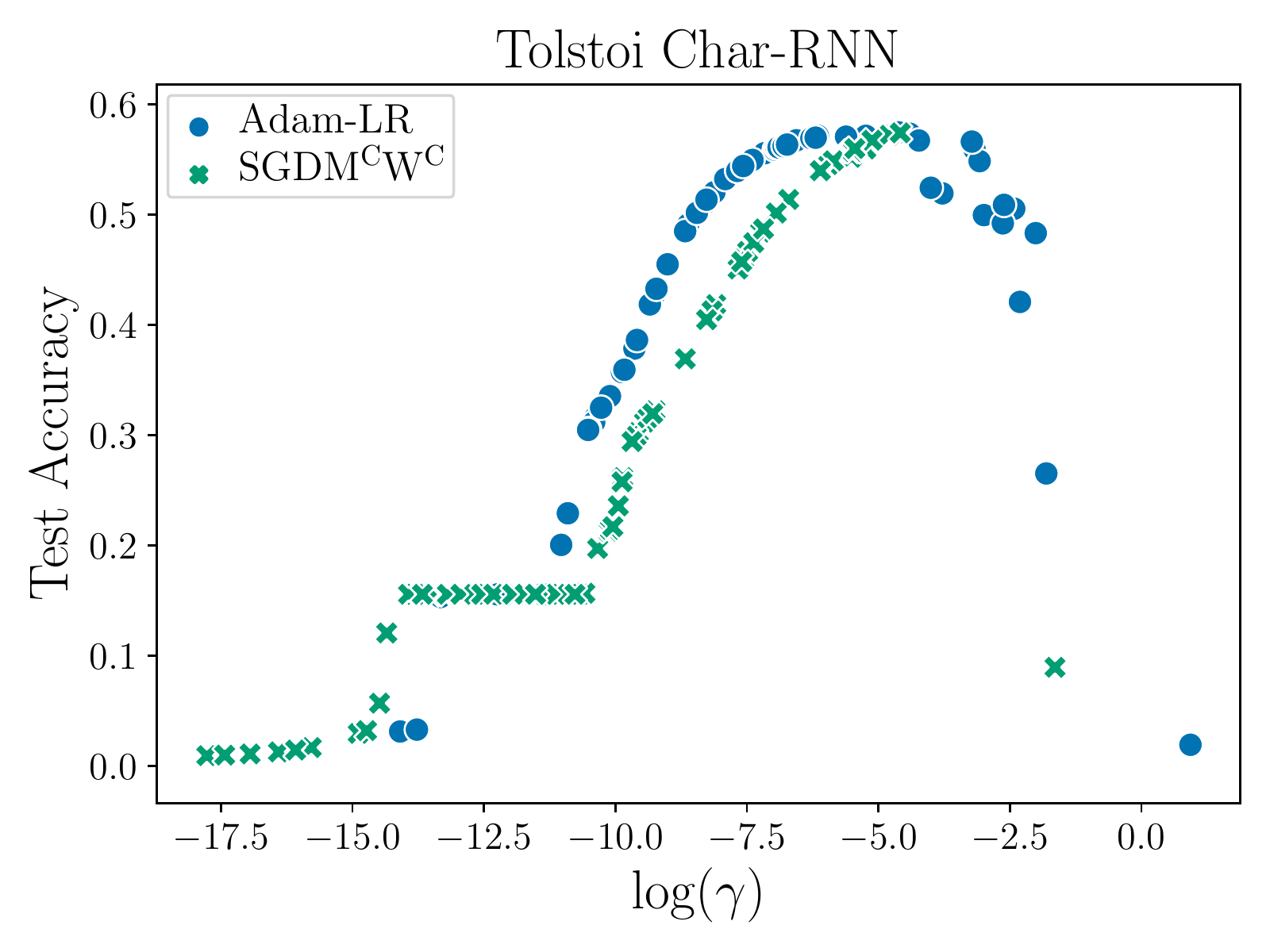}
    \end{subfigure}
    
    \caption{Scatter plot of performance of \AdamLR and \SGDMCWC by learning rate value. For better visibility, we shift the learning rate values of \SGDMCWC in such a way that the minima of both optimizers are at the same position on the x-axis.}
    \label{fig:hyperparam_surface}
\end{figure*}

In Figure~\ref{fig:hyperparam_surface}, we show a scatter plot of the loss/accuracy surfaces of \SGDMCWC and \AdamLR as a function of the learning rate, which is their only tunable hyperparameter. The plots confirm the expected behavior. On MNIST VAE, FMNIST VAE, and Tolstoi-Char-RNN, \AdamLR reaches performances close to the optimum on a wider range of learning rates than \SGDMCWC does, resulting in substantially better expected performances at small budgets ($k = 1,4$) as opposed to \SGDMCWC, even though their extrema are relatively close to each other. On CIFAR10, the width of the maximum is similar, leading to comparable performances at low budgets. However, the maximum for \SGDMCWC is slightly higher, leading to better performance than \AdamLR at high budgets. 

\section{Interplay between momentum and learning rate}\label{app:sec:lreff}
We ran an additional experiment using `effective learning rate'~\citep{shallue2019measuring} that combines learning rate $\gamma$, and momentum $\mu$ of SGD to compute the effective learning rate \lreff. Intuitively, \lreff quantifies the contribution of a given minibatch to the overall training. This is defined as \[\lreff = \frac{\gamma}{1-\mu}\]

\begin{figure*}[htp]
    \begin{subfigure}[t]{0.33\textwidth}
        \includegraphics[width=\textwidth]{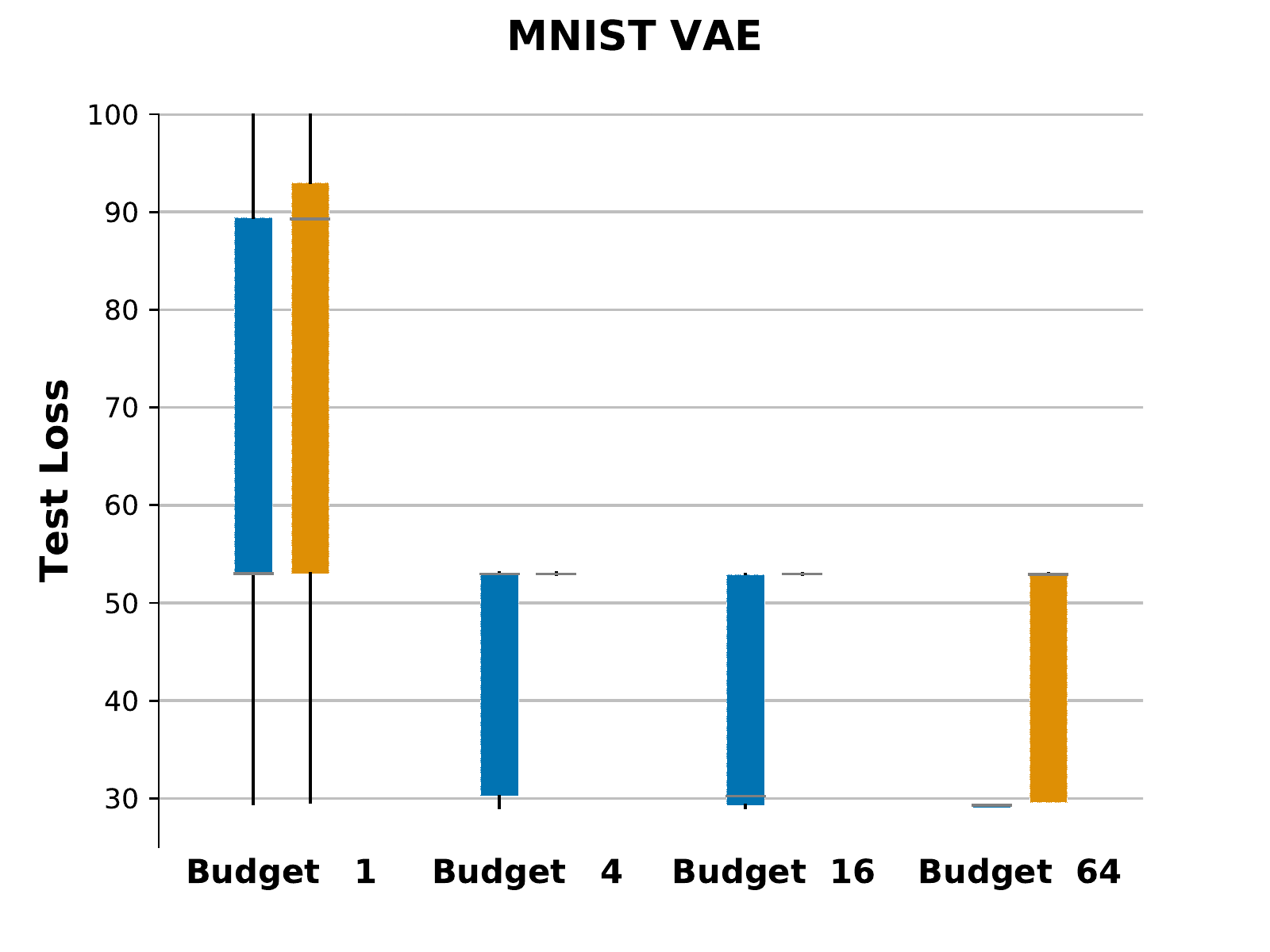}
    \end{subfigure}
    \begin{subfigure}[t]{0.33\textwidth}
        \includegraphics[width=\textwidth]{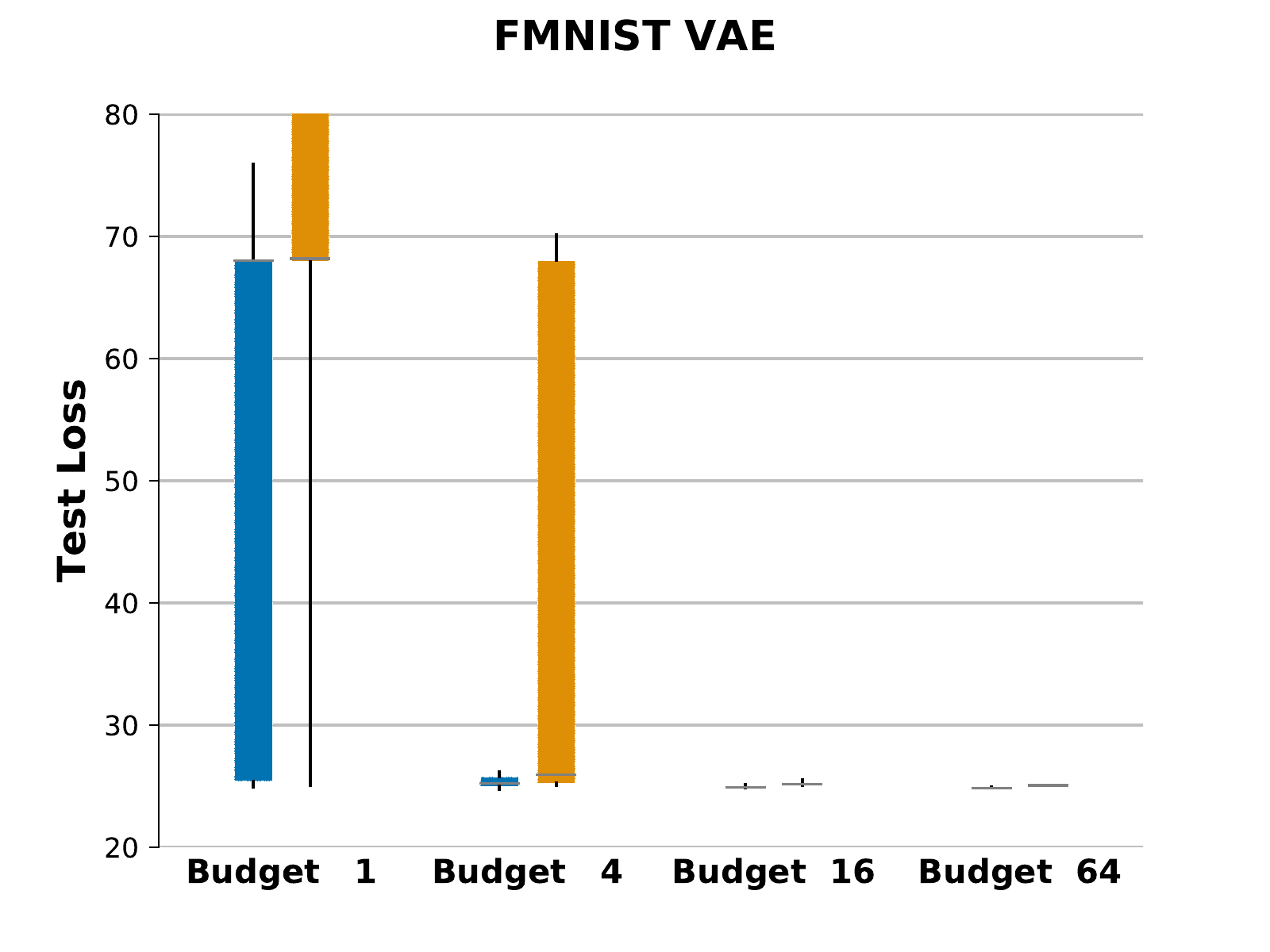}
    \end{subfigure}
    \begin{subfigure}[t]{0.33\textwidth}
        \includegraphics[width=\textwidth]{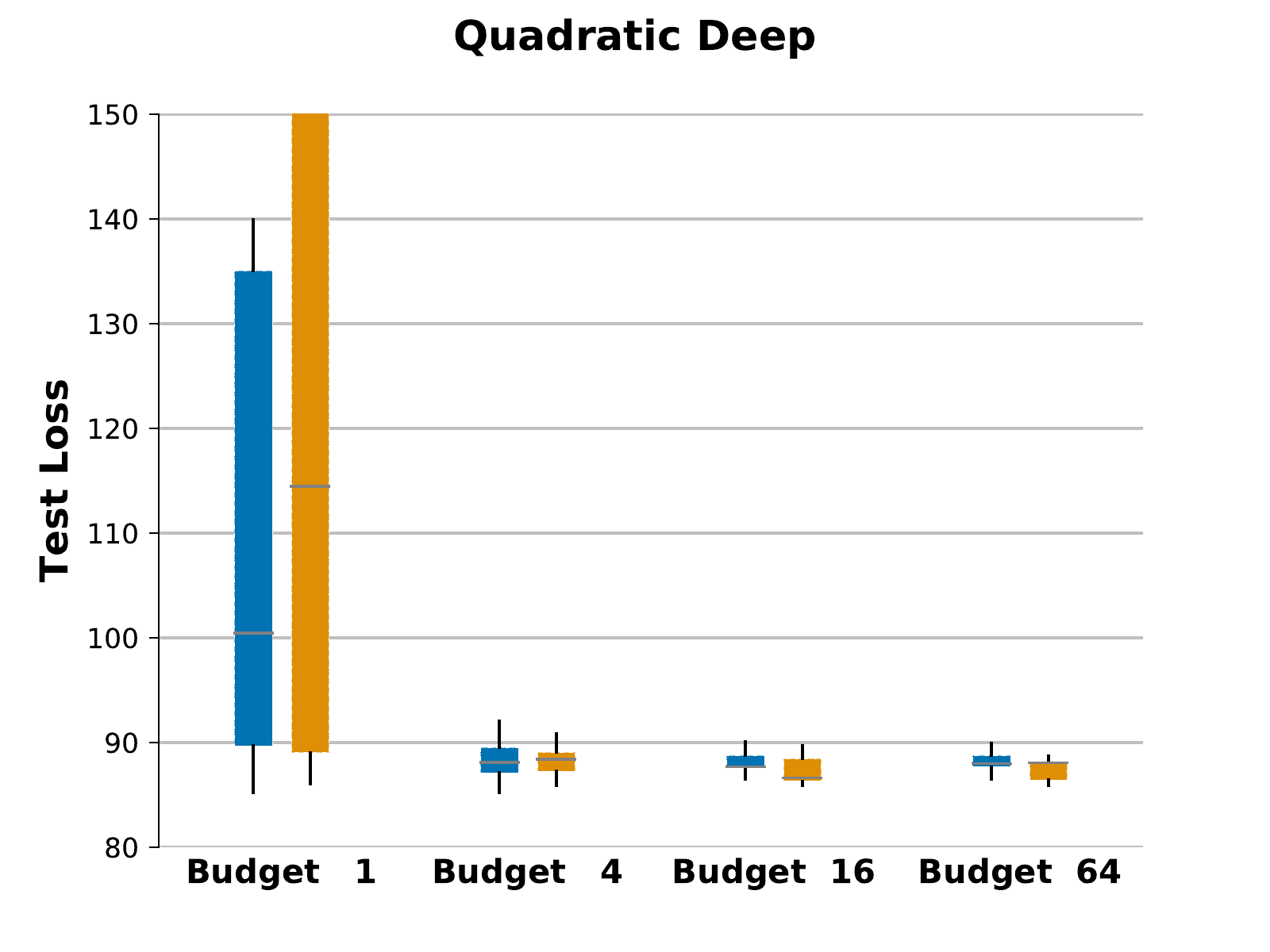}
    \end{subfigure}
    \begin{subfigure}[t]{0.33\textwidth}
        \includegraphics[width=\textwidth]{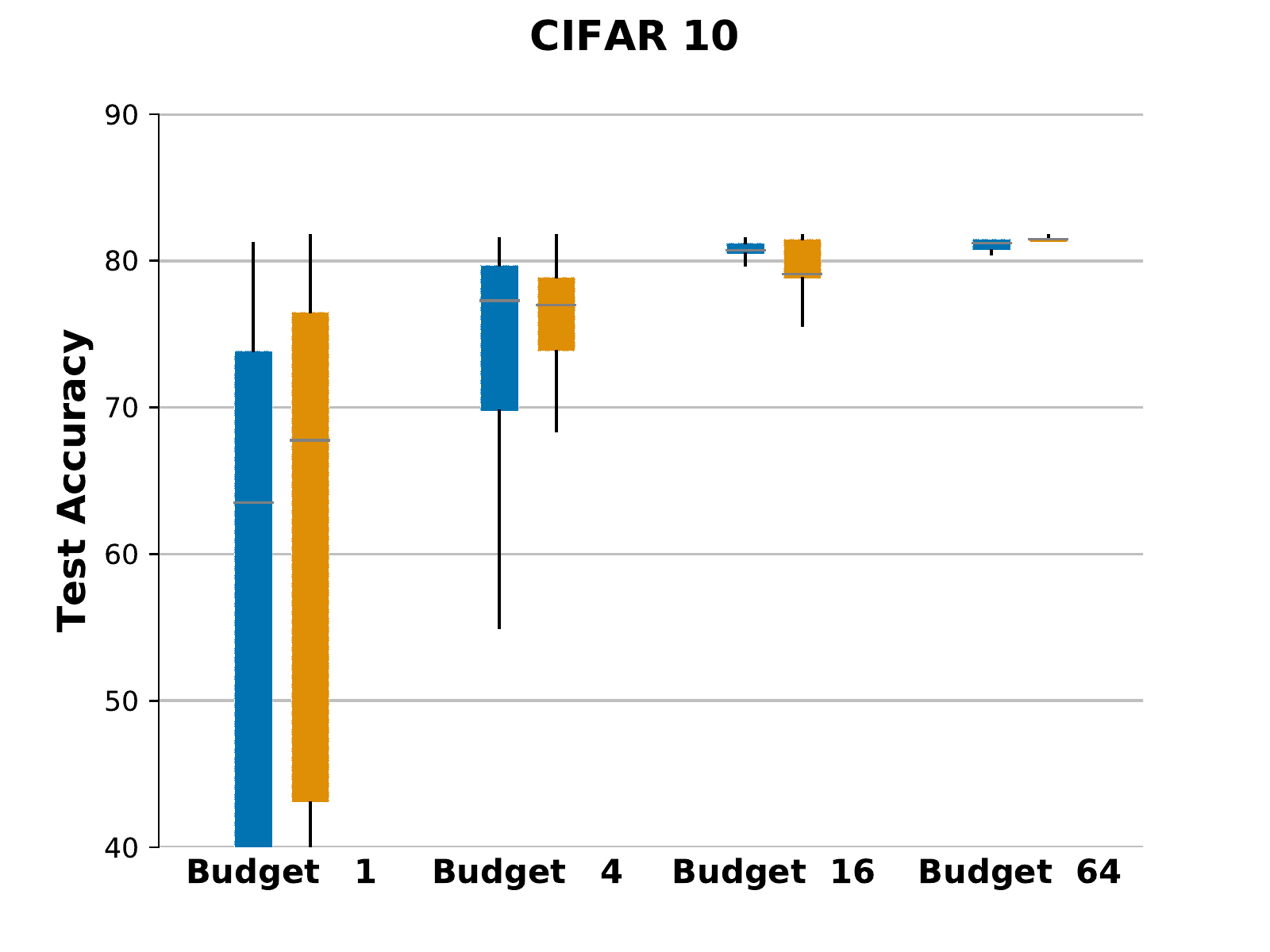}
    \end{subfigure}
    \begin{subfigure}[t]{0.33\textwidth}
        \includegraphics[width=\textwidth]{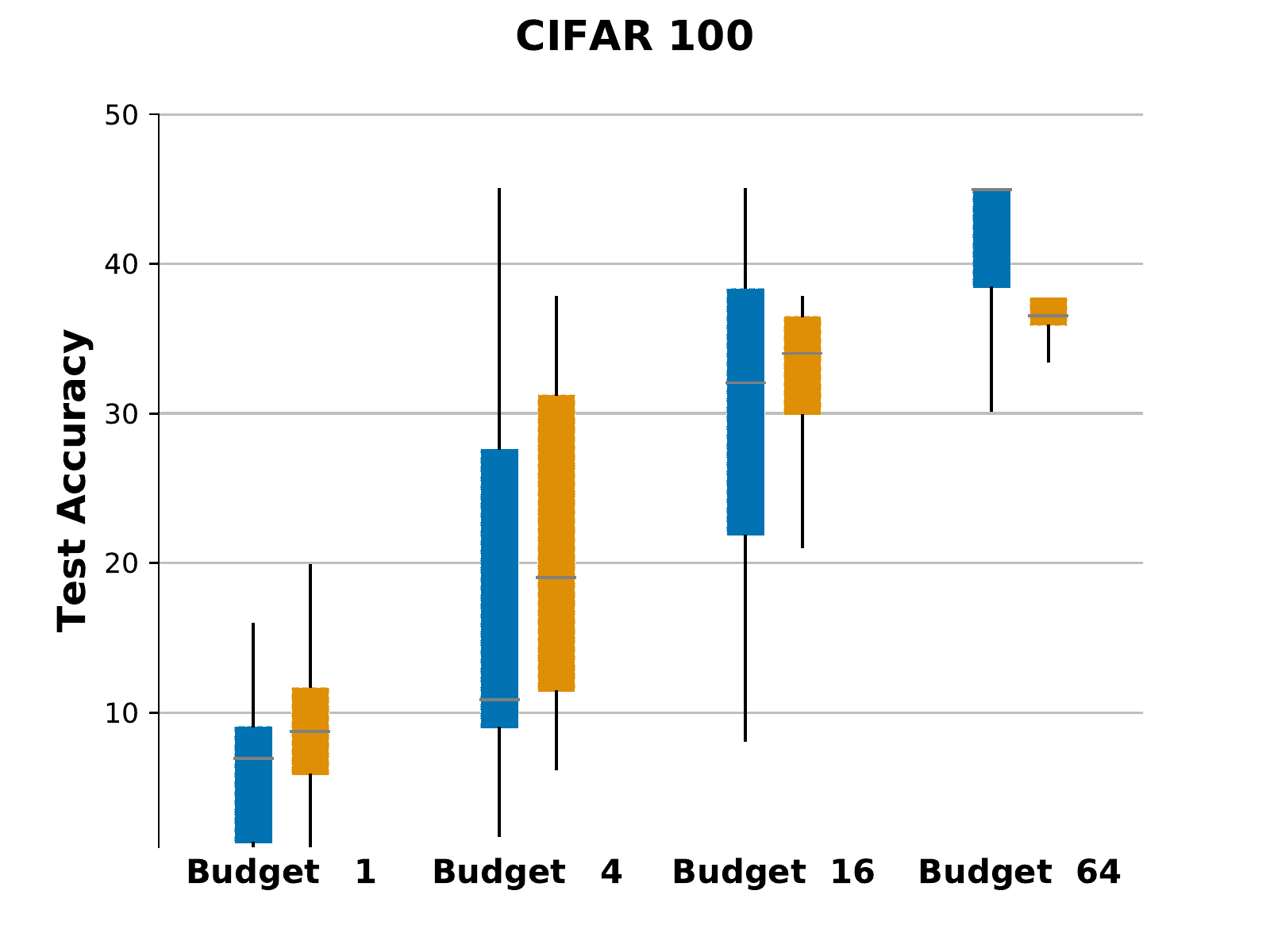}
    \end{subfigure}
    \begin{subfigure}[t]{0.33\textwidth}
        \includegraphics[width=\textwidth]{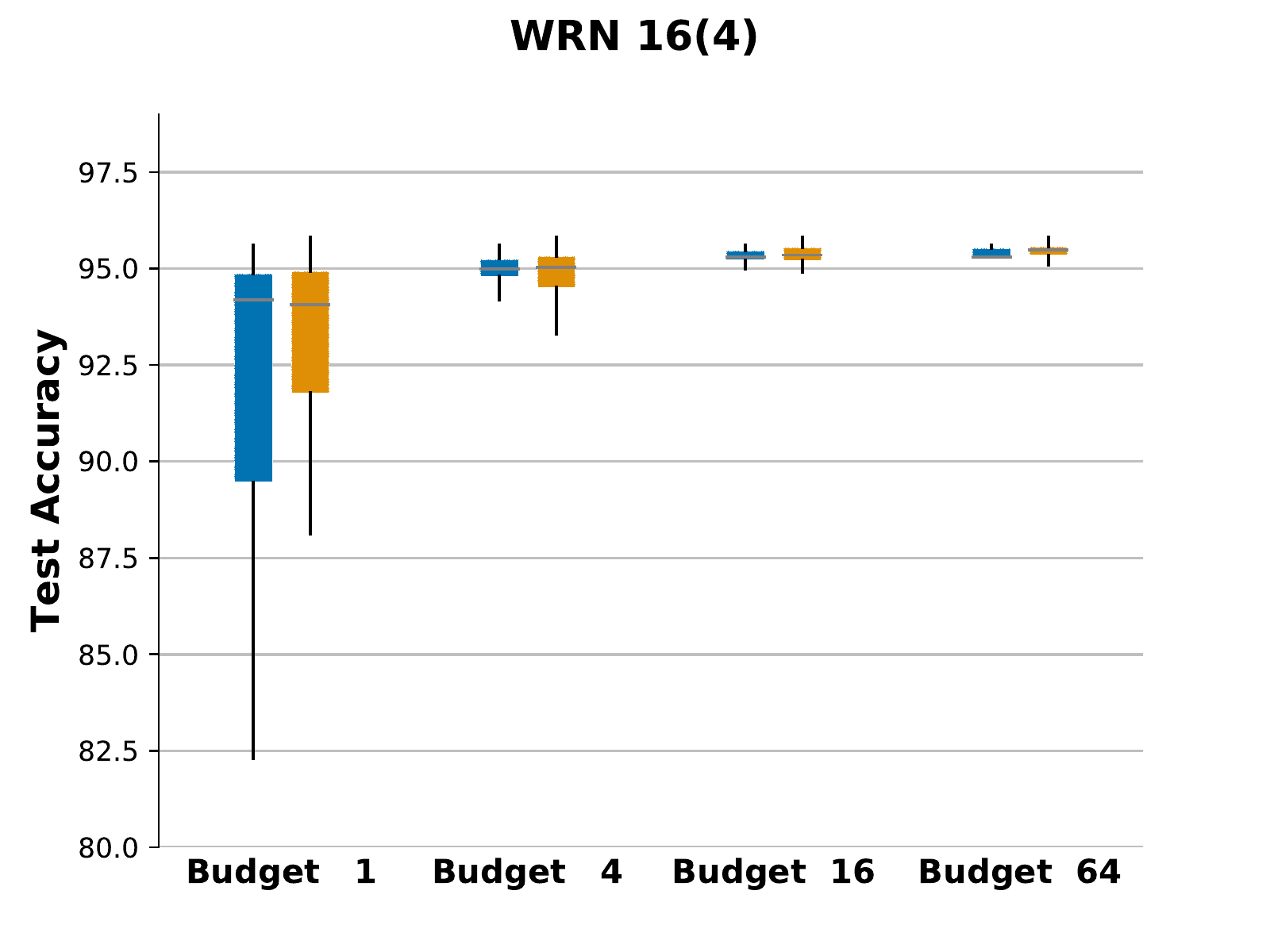}
    \end{subfigure}
    \begin{subfigure}[t]{0.33\textwidth}
        \includegraphics[width=\textwidth]{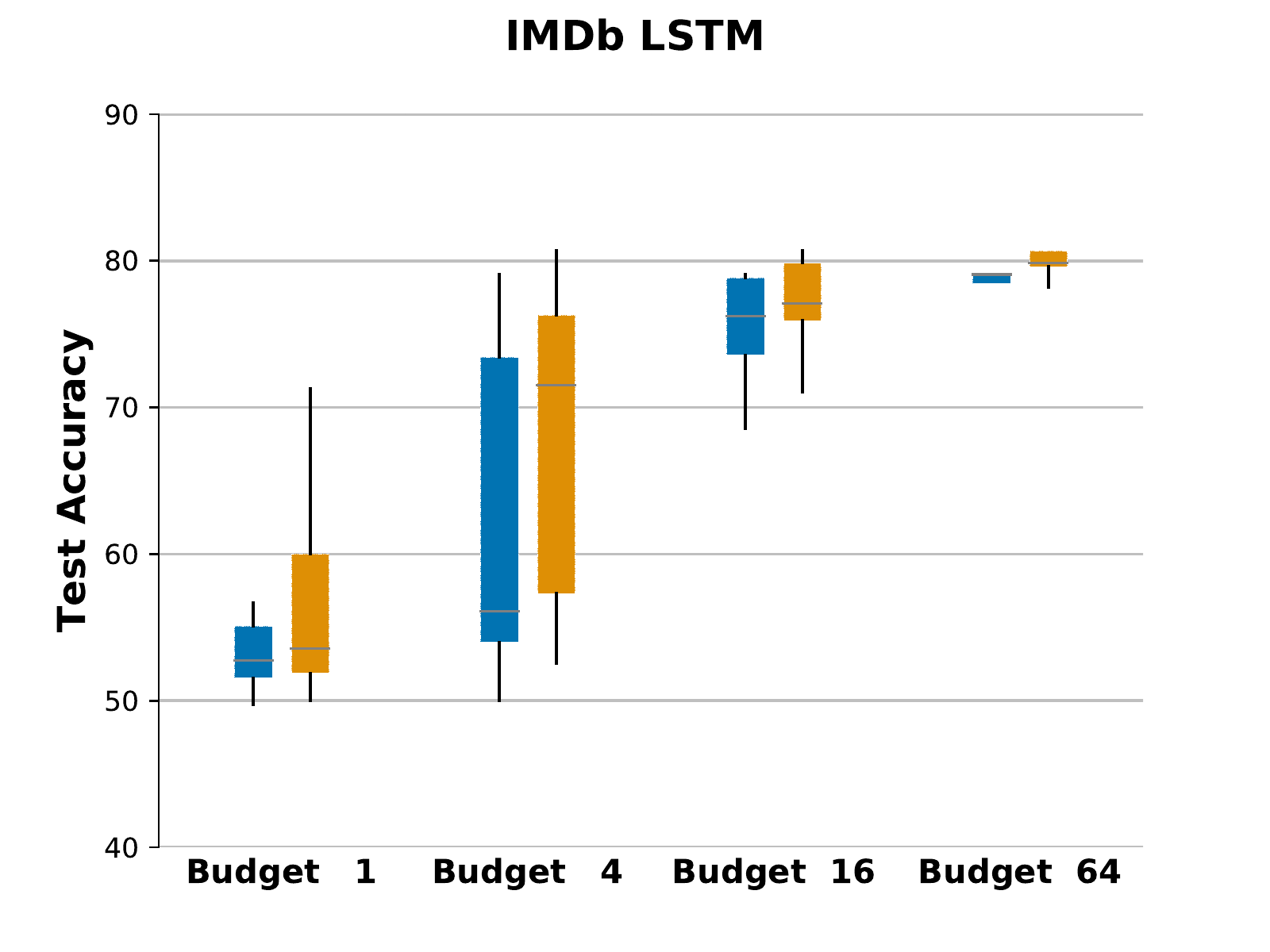}
    \end{subfigure}
    \begin{subfigure}[t]{0.33\textwidth}
        \includegraphics[width=\textwidth]{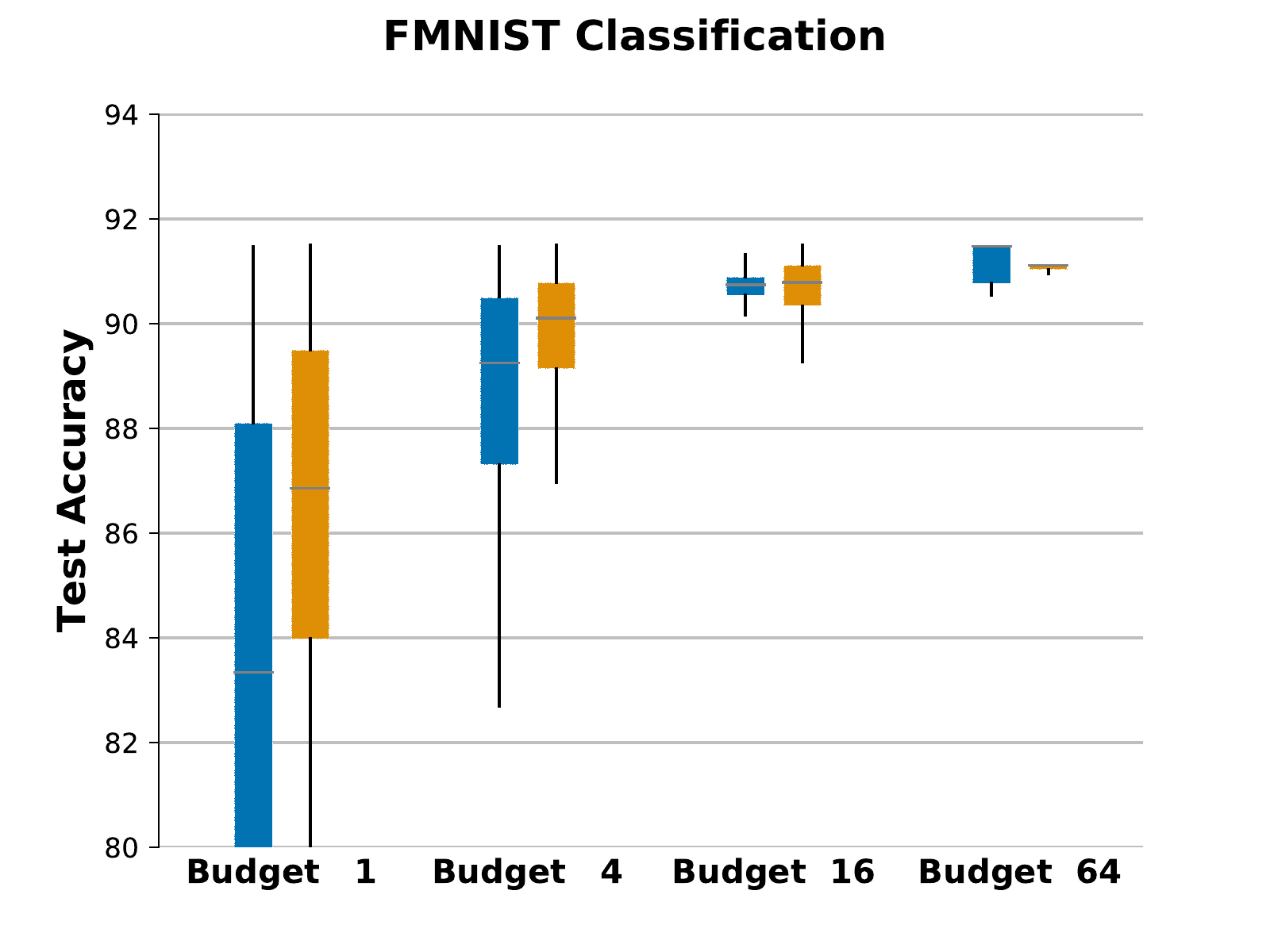}
    \end{subfigure}
    \begin{subfigure}[t]{0.33\textwidth}
        \includegraphics[width=\textwidth]{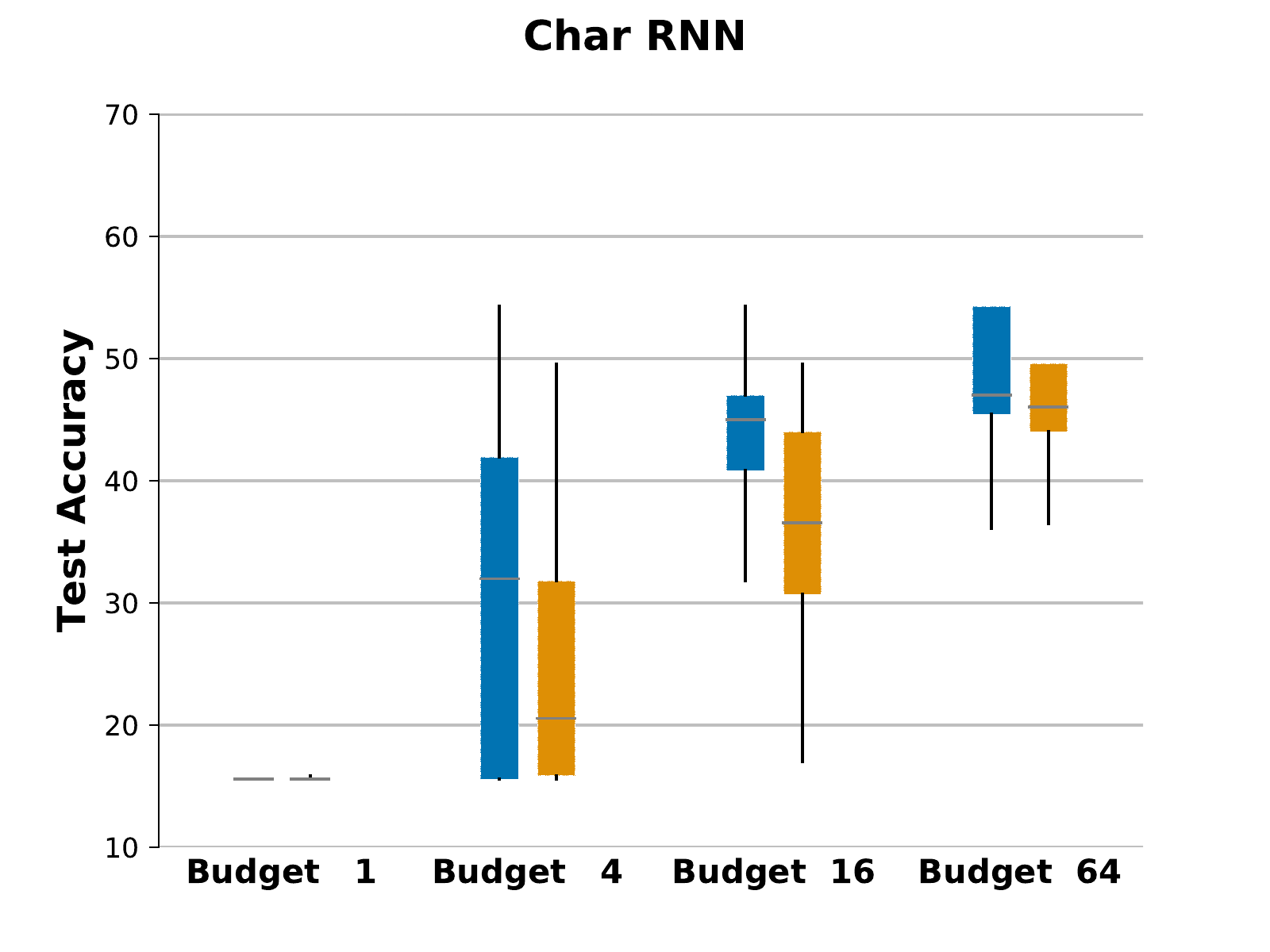}
    \end{subfigure}
    
    \caption{Performance of \showcolor{color1}{\SGDMW}, \showcolor{color2}{\SGDLreff}, at various hyperparameter search budgets. Image is best viewed in color. Some of the plots have been truncated to increase readability.}
    \label{fig:lreff_plots}
\end{figure*}

\begin{figure*}[htpb]
  \begin{center}
    \includegraphics[width=0.40\textwidth]{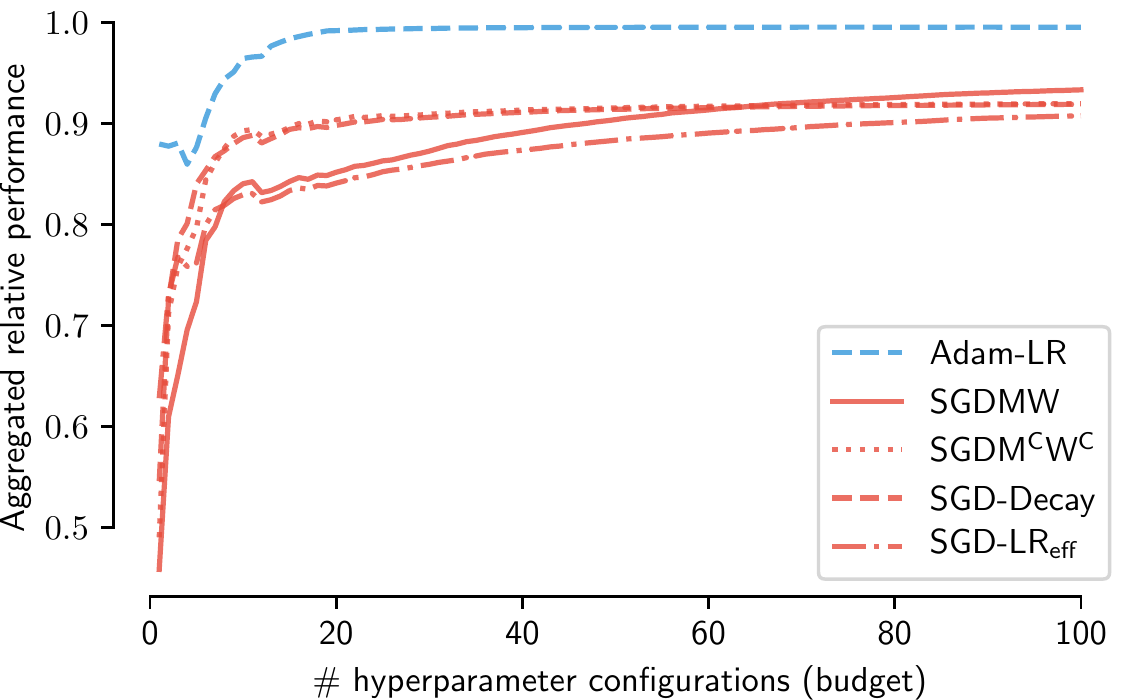}
  \end{center}
  \vspace{-0.5\baselineskip}
  \caption{
    Aggregated relative performance of \SGDLreff compared to other optimizers.
  }
  \label{fig:lreff_summary}
\end{figure*}

We designed a variant of \SGDMW, called \SGDLreff, where we sampled $\gamma$ and \lreff independently from lognormal priors calibrated as usual, and compute the momentum($\mu$) as $\mu = \max(0, (1-\frac{\gamma}{\lreff}))$, hence accounting for interplay between learning rate and momentum. We plot the performance comparisons between \SGDMW and \SGDLreff in \cref{fig:lreff_plots}, and provide a plot of the aggregated relative performance in \cref{fig:lreff_summary}. 
The results show that indeed \SGDLreff improves over \SGDMW in the low-budget regime, particularly on classification tasks. We attribute this to the fact that \SGDLreff is effective at  exploiting historically successful $(\gamma, \mu)$ pairs. For large budgets, however, \SGDLreff performs increasingly worse than \SGDMW, which can be explained by the fact that \SGDMW has a higher chance of exploring new configurations due to the independence assumption. Despite the improvement in low-budget regimes, SGD variants, including the new \SGDLreff variant, remain substantially below \AdamLR in all budget scenarios. Hence, our conclusion remains the same.

\end{appendices}
\end{document}